\documentclass[conference]{IEEEtran}
\IEEEoverridecommandlockouts
\usepackage{cite}
\usepackage{amsmath,amssymb,amsfonts}
\usepackage{algorithmic}
\usepackage{graphicx}
\usepackage{textcomp}
\usepackage{xcolor}
\usepackage{subcaption}
\usepackage{hyperref}
\usepackage{diagbox}

\def\BibTeX{{\rm B\kern-.05em{\sc i\kern-.025em b}\kern-.08em
    T\kern-.1667em\lower.7ex\hbox{E}\kern-.125emX}}
\begin{document}

\title{Grasping Force Estimation for Markerless
Visuotactile Sensors\\

\thanks{This work was supported  by Interreg-VI Sudoe and European Regional Development Fund through the REMAIN project (S1/1.1/E0111) and by the University of Alicante under Grant UAFPU21-26.}
\thanks{Julio Castaño-Amoros is with the AUROVA Lab, Computer Science Research
Institute, University of Alicante, 03690 Alicante, Spain 
        {\tt\small julio.ca@ua.es}}
\thanks{Pablo Gil is with the AUROVA Lab, Computer Science Research Institute,
University of Alicante, 03690 Alicante, Spain, and also with the Department
of Physics, Systems Engineering, and Signal Theory, University of Alicante, 03690 Alicante, Spain
        {\tt\small pablo.gil@ua.es}}
}

\author{\IEEEauthorblockN{1\textsuperscript{st} Julio Castaño-Amoros}
\and
\IEEEauthorblockN{2\textsuperscript{nd} Pablo Gil}
}

\maketitle

\begin{abstract}
Tactile sensors have been used for force estimation in the past, especially Vision-Based Tactile Sensors (VBTS) have recently become a new trend due to their high spatial resolution and low cost. In this work, we have designed and implemented several approaches to estimate the normal grasping force using different types of markerless visuotactile representations obtained from VBTS.
Our main goal is to determine the most appropriate visuotactile representation, based on a performance analysis during robotic grasping tasks.
Our proposal has been tested on the dataset generated with our DIGIT sensors and another one obtained using GelSight Mini sensors from another state-of-the-art work. We have also tested the generalization capabilities of our best approach, called \textit{RGBmod}. The results led to two main conclusions. First, the RGB visuotactile representation is a better input option than the depth image or a combination of the two for 
estimating normal grasping forces. Second, \textit{RGBmod} achieved a good performance when tested on 10 unseen everyday objects in real-world scenarios, achieving an average relative error of 0.125 $\pm$ 0.153. Furthermore, we show that our proposal outperforms other works in the literature that use RGB and depth information for the same task.
\end{abstract}

\begin{IEEEkeywords}
Tactile sensing, Force prediction, Grasping Force\end{IEEEkeywords}

\section{Introduction}

\label{sec:introduction}
\label{sec:intro}
Humans can process a large amount of tactile information in order to perform various manipulation tasks. This information includes temperature, weight, rigidity, texture, slippage, and fragility. The degree to which we rely on this information depends on our intuition and prior experience with the force applied during grasping and manipulation tasks.

Although it is not possible to accurately estimate the applied force in newtons (N), we have learned from experience to use our sense of sight to estimate a force prior to grasp an object.
However, we still need our sense of touch in order to adjust the applied force if the force prior is inaccurate. A clear example of this is a bottle filled with a liquid that is not visible from the outside. In this case, we can identify the material of which the bottle is made by sight and estimate the force required in order to grasp it.
However, we must also adjust the grasping force so as to account for the weight of the liquid. 
This example can be applied to the field of robotic manipulation, where the goal is to achieve human-like dexterity \cite{trends_and_challenges}. Computer vision can, therefore, be used in robotic manipulation tasks to estimate force priors, but the use of tactile sensors is also necessary in order to obtain real-time force feedback and adjust the grasping force as needed.

Various tactile sensors can be used to estimate force values, such as resistive, capacitive, magnetic, optical, visuotactile sensors, etc.
Note that those sensors do not directly provide a force magnitude in N, but that it is usually obtained from the measurement of other related physical magnitudes.

For example, in \cite{capacitive_sensor} the authors used a capacitive tactile sensor to obtain the normal 
and the tangential forces to the surface of the sensor. They obtained four capacitance values by applying the parallel electrode capacitor equations, and then mapped them to force values using linear equations. 
In addition, magnetic-based tactile sensors measure the magnetic flux caused by the deformation of the magnetic elastic layer of the sensor.
In \cite{magnetic_sensor}, this magnetic flux is related to the normal and tangential forces applied to the surface of the sensor using non-linear equations.

Other types of tactile sensors rely
on optical instruments such as photodiodes (optical tactile sensors) or cameras (VBTS) \cite{review}. 
Photodiodes are used to measure the change in illumination in volts caused by the deformation of the sensor surface. For example, the authors of \cite{papilarray} used a 5th order multivariate polynomial regression model to map voltage values to normal and tangential force values. In contrast, the VBTS discussed in the following section use a camera (usually RGB) to record the deformation on the gel of the sensor.
All of the above sensors can be used to estimate forces from their respective raw signals. 
However, capacitive, resistive, magnetic and optical sensors have similar limitations compared to VBTS. These limitations are a low spatial resolution and a high cost.
High spatial resolution allows VBTS to capture fine-grained features such as textures, shapes, contours, depth, etc, of the contact surface. Moreover, VBTS are very economical, making them more accessible to the industry and academia.
For these reasons, we have decided to use VBTS like DIGIT \cite{digit} sensors to estimate grasping forces from tactile images. 

The contribution of our work is threefold:

\begin{itemize}
    \item We propose a methodology for estimating grasping forces from markerless tactile images based on the ResNet-18 architecture. We have tested our proposal with different sensors and datasets, obtaining generalized performance and robust results in real-world scenarios. Moreover, in this paper, we have proved that our best method outperforms other state-of-the-art works that do not use visual markers.

    \item We have performed an exhaustive analysis of the use of different visuotactile representations (RGB, depth, mixed) as input, proving that the proposed method is more accurate when using the RGB data representation with the DIGIT sensors. We have also investigated the limitations of using depth tactile images for force estimation.

    \item We have collected a dataset related to the force estimation task from the tactile data of the DIGIT sensors. Our dataset includes RGB tactile images, depth tactile images, and ground truth forces. We believe that our dataset contributes to the literature of the proposed task.
\end{itemize}

The remainder of the paper is organized as follows: Section \ref{sec:related_work} describes different force estimation methods from the literature that use VBTS, while Section \ref{sec:methods} explains the visuotactile representations and the proposed methods. Section \ref{sec:experimental_setup} describes our robotic setup and the collection process of our dataset. Section \ref{sec:experimental_results_analysis} shows the results obtained after grasping objects in different experiments: on our dataset (indenters), on unseen everyday objects with symmetrical and non-symmetrical contact surfaces, and after comparing our proposal with other state-of-the-art works. Our work concludes in Section \ref{sec:conclusions}, which provides a summary of our proposal with emphasis on the most significant results, the limitations and the future work.

\section{Background}

\label{sec:related_work}
According to the state of the art, several approaches can be considered to try to solve the problem of estimating the force of touch when using VBTS. We have decided to classify them into two categories depending on whether the visuotactile representation contains visual markers or not. 

 Regarding the works in the literature that used visual markers, there seem to be two predominant methodologies: the use of nonlinear models such as neural networks \cite{sferraza1}, \cite{sferraza2}, \cite{yuan}, \cite{helmholtz}, \cite{cross-modal}, \cite{gelstereo}, \cite{msresnet}, and the estimation of the calibration matrix that represents the deformation model of the visuotactile sensor  \cite{alberto}, \cite{althoefer}.

In \cite{sferraza1}, the authors mapped the displacement of the visual markers to a 3D force distribution using optical flow and a deep neural network. They later extended their work in \cite{sferraza2} by integrating a simulator into their pipeline, which allowed them to train a UNET-like Convolutional Neural Network (CNN) to obtain the 3D force distribution from simulated tactile images.
In \cite{yuan}, a binarized tactile image with visual markers was used as input to a VGG-like CNN to obtain normal and shear forces and a single torque value. In \cite{helmholtz}, the authors trained a Multilayer Perceptron (MLP) to estimate a normal force, a shear force, and a torque value from the Helmholtz-Hodge Decomposition of the displacement of the markers. Similarly, \cite{cross-modal} used an MLP to estimate 3D normal and shear forces from the displacement of the markers using their custom sensor.
The authors of \cite{gelstereo} implemented a neural network architecture to map from 2D and 3D marker displacements to 6D force and torque values using the GelStereo tactile sensor. In \cite{msresnet}, the authors compared state-of-the-art CNNs to compute 3D forces and torques from tactile images with visual markers.

In addition, other authors in \cite{alberto} and \cite{althoefer} estimated the sensor-specific calibration matrix, using a Finite Element Analysis and a calibration process with a force/torque sensor, respectively. The forces were obtained by multiplying the displacements of the visual markers and the calibration matrix. A 3D force distribution was obtained in the case of \cite{alberto}, while 6D force and torque values were estimated in \cite{althoefer}.

Most of the work related to force regression from visuotactile data has used visual markers, while other types of visuotactile representations that do not use these markers, such as RGB images \cite{2}, \cite{11}, \cite{minsight}, \cite{digit_pinki}, \cite{9dtact}, depth data \cite{shukrullo}, or a combination of both, have not been explored as thoroughly. Although the use of markers facilitates the estimation of the grasping force, adding markers to a visuotactile sensor makes the manufacturing process more expensive. Moreover, preprocessing the markers can also increase the computational time when compared to that of an end-to-end method such as the one presented in this paper.

In \cite{2}, the authors used a GelSight Mini sensor and an end-to-end UNET-like CNN to estimate a single normal force value and to segment the contact region between the object and the sensor. They then combined these outputs by dividing the normal force by the area of the contact region to obtain a normal force distribution containing very similar forces within the segmented region. In \cite{11}, other authors presented Insight, a soft thumb-sized VBTS that provides a 360º RGB tactile image. They also used CNNs to train a ResNet model to map the tactile image to a 3D force map. However, the Insight sensor is too large to fit on a robotic gripper or hand, so a miniaturized version, called Minsight was presented in \cite{minsight} to be used for this task. In this line, the authors of \cite{digit_pinki} presented a new miniaturized version of the DIGIT sensor, called DIGIT Pinki. This new version was constructed using fiber optic bundles to reduce the size of the sensor while keeping the electronics and camera remote. The authors trained a ResNet-18 model to map the tactile images to normal forces (up to 1N). In addition, a DIGIT sensor was calibrated in \cite{shukrullo} to map its RGB images to depth images using a spherical indenter.
The authors later mapped depth data to a single normal force value by fitting a polynomial regression model, which was tested in a teleoperation task with a cylindrical object. In \cite{9dtact}, the authors presented a new VBTS, called 9DTact, with which they could obtain 3D reconstruction, 3D force and torque values. In this work, they trained the DenseNet CNN architecture to map a preprocessed tactile image to forces and torques.

In this work, we chose to base our proposal on approaches that do not use visual markers. We believe that these markers may cause a loss of tactile information due to occlusions (e.g., textures, contact shapes, etc.), that can be useful for robotic manipulation tasks. In addition, there is less research in the literature on the force regression task without visual markers, focusing primarily on normal force regression with DIGIT-like sensors. Furthermore, there is no clear guidance on the most appropriate type of visuotactile or tactile data representation for this specific task. For this reason, we have decided to study it in this paper.

Specifically, our work differs from others that do not use markers in several ways. For example, \cite{2} approached the task of force regression as image segmentation with pixel-wise regression. In contrast, our approach is based on estimating only a single force value for the tactile image, which reduces the complexity of the task. Although one might think that pixel-wise regression provides a higher spatial resolution (a total of 76,800 values) than a single regression value, the force distribution estimated in \cite{2} contained almost identical force values because they used a single ground truth force.

In other works, such as \cite{11}, \cite{minsight}, and \cite{digit_pinki}, their authors used omnidirectional VBTS, which may be more suitable for use in robotic hands. However, we work with parallel grippers that apply forces only on the front side of the sensor, while robotic hands can apply forces on different sides of the fingertips. Therefore, there is no advantage to using omnidirectional sensors instead of DIGIT sensors.

Furthermore, DIGIT can provide features such as lights and shadows from the captured images without additional preprocessing to obtain a tactile image, which enhances these features, as done in \cite{9dtact}.
In contrast to \cite{shukrullo}, where the force was estimated from a single depth value obtained as the maximum deformation of the sensor surface, we have used not only depth but also RGB data.

Therefore, in this paper, we aim to perform rigorous and extensive experiments comparing the aforementioned visuotactile representations used in our proposed methods for grasping force estimation using a DIGIT sensor without visual markers. Consequently, this paper focuses on comparing our work with other works in the literature that use similar VBTSs (DIGIT-like) without visual markers, such as \cite{2} and \cite{shukrullo}. Note that the task of performing force regression using VBTS without visual markers is challenging. Therefore, it would be inappropriate to directly compare methods that do not use visual markers with those that do.

\section{Methods}
\label{sec:methods}

\subsection{Visuotactile Representations}
\label{sec:calib_process}
As mentioned above, the goal of this work is to estimate the grasping normal force using the tactile data from the DIGITs as input and the forces obtained from the PapillArray tactile sensor \cite{papilarray} as the ground truth. Two different handmade DIGIT sensor units are used, capturing 320x240x3 RGB images at 60 FPS, while a single PapillArray sensor is used in order to obtain the ground truth global force at a frequency of 500Hz. The data from both sensors are synchronized to the frequency of the DIGIT sensors using Robot Operating System (ROS).

In addition to using the RGB images captured by the DIGIT sensors, we generate depth images that represent the physical deformation that the sensor gel undergoes during grasping. The process used to generate the depth images is based on \cite{yuan} and \cite{song-thesis}.

Photometric stereo has often been used to estimate the surface normals of objects by observing them under different illuminations. The deformation of the gel caused by contact with an object changes the amount of light reflected from its surface, thus affecting the captured RGB image. Therefore, changes in deformation cause changes in the colors of the tactile image. Our objective is to find a mathematical model that maps the color and position of pixels to depth values in order to obtain our visuotactile representation based on the depth image. The procedure is as follows.

On the one hand, given a depth map defined as $z=H(x,y)$, where $(x,y,z)$ represent the Euclidean coordinates of each pixel, the surface normal $N(x,y,z)$ can be calculated as depth gradients by deriving $H$ with respect to each one of the directions as shown in Eq. \ref{eq:theory_normals}.

\begin{equation}
    N(x,y,z) = (\frac{\partial{H}}{\partial{x}}, \frac{\partial{H}}{\partial{y}},1)
    \label{eq:theory_normals}
\end{equation}

On the other hand, we can relate each pixel $i_{xy}=\{r,g,b\} \in$ the color image $I(x,y)$ to the surface normal by estimating a nonlinear function called $R$ which is shown in Eq. \ref{eq:color_intensity_from_normals}. We will need to determine the inverse function of $R$ to map from color intensity to depth gradients.

\begin{equation}
    I(x,y) = R(\frac{\partial{H}}{\partial{x}} + \frac{\partial{H}}{\partial{y}})
    \label{eq:color_intensity_from_normals}
\end{equation}

Finally, the Poisson Equation (see Eq. \ref{eq:poisson_eq}) is solved to integrate the depth gradients in order to obtain the depth map.

\begin{equation}
    \nabla^{2}H = \frac{\partial^2{H}}{\partial{x}^2} + \frac{\partial^2{H}}{\partial{y}^2}
    \label{eq:poisson_eq}
\end{equation}

where $\nabla^{2}$ refers to the Laplace operator.

Considering the above, the first step is to approximate the $R^{-1}$ function to map from color intensity $\{r,g,b\}$ to normal vectors $\{n_{x}, n_{y}, n_{z}\}$ for each pixel $i_{xy} \in I(x,y)$.
To do this, we collect a small dataset (40 images) by pressing a spherical object of known radius at different positions on the surface of the sensor. Then, a normal vector is calculated for each pixel using the spherical coordinate system (see Eq. \ref{eq:surface_normals}).

\begin{equation}
\begin{gathered}
    n_{x} = cos(\phi) * cos(\theta) \\
    n_{y} = cos(\phi) * sin(\theta) \\
    n_{z} = sin(\theta)
    \label{eq:surface_normals}
\end{gathered}
\end{equation}

where $\theta$ is the azimuth angle and $\phi$ is the elevation angle of the sphere. An MLP neural network is then trained to approximate the $R^{-1}$ function using the collected dataset. Therefore, we compute the depth gradients from the normal vectors according to Eq. \ref{eq:theory_normals} by normalizing the normal values as shown in Eq. \ref{eq:gradients}.

\begin{equation}
     (\frac{\partial{H}}{\partial{x}}, \frac{\partial{H}}{\partial{y}},1) = \{\frac{n_{x}}{n_{z}}, \frac{n_{y}}{n_{z}}, \frac{n_{z}}{n_{z}}\}
    \label{eq:gradients}
\end{equation}

\begin{figure}[htbp]
    \centering
\includegraphics[scale=0.45]{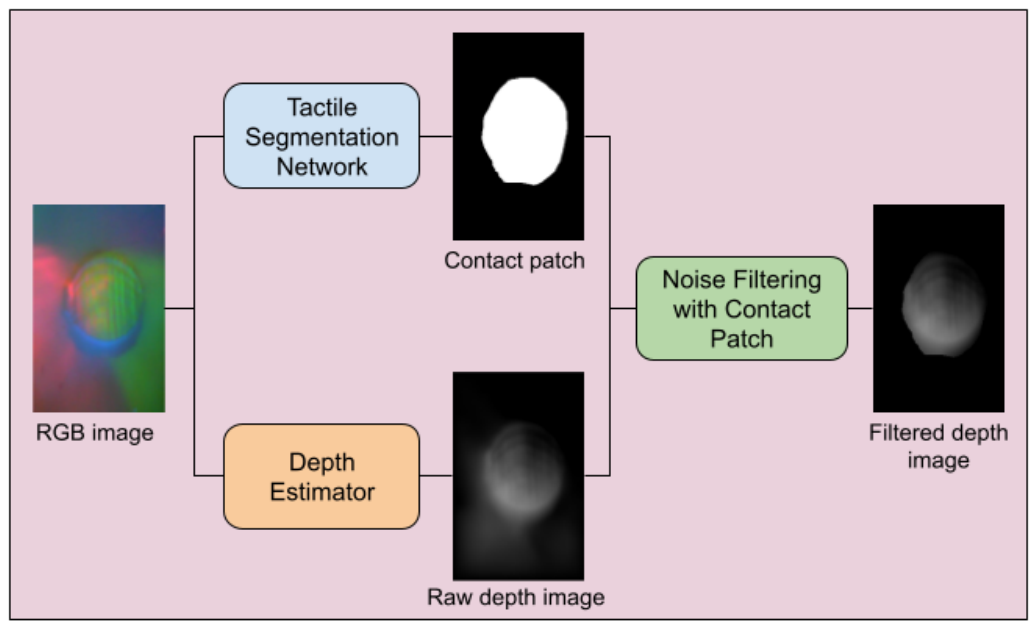}
    \caption{Depth estimation with noise filtering from the RGB image}
    \label{fig:depth_filt_mask}
\end{figure}

Finally, the depth map is obtained by solving Eq. \ref{eq:poisson_eq} using a fast Poisson Solver with a Discrete Sine Transform (DST). In our work, we scale this depth map in order to transform it into a depth image with dimensions of 320x240x1, whose pixel values $d_{xy}$ are between 0 and 255. The depth image contains background noise caused by inaccuracies during the calibration process, as shown in Fig \ref{fig:depth_filt_mask}. This noise is filtered by using our tactile segmentation network \cite{julio-pablo-ral} to obtain the binary contact patch from the RGB image, and then applying this patch as a mask over the depth image.

\begin{figure}[htbp]
     \centering
     \begin{subfigure}[b]{0.1\textwidth}
         \centering
         \includegraphics[scale=0.28]{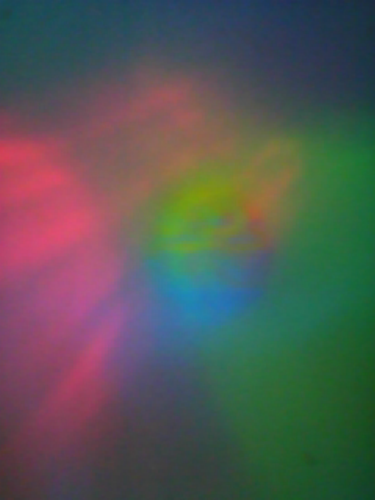}
         \caption{}
     \end{subfigure}
     \begin{subfigure}[b]{0.1\textwidth}
         \centering
         \includegraphics[scale=0.28]{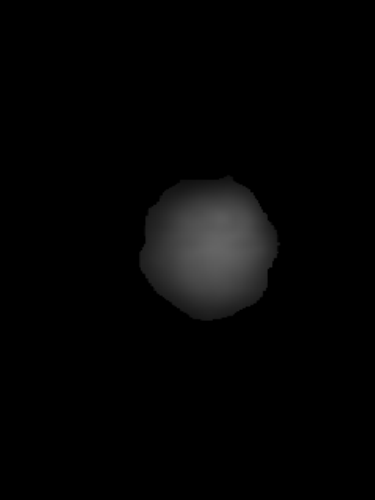}
         \caption{}
     \end{subfigure}
     \begin{subfigure}[b]{0.1\textwidth}
         \centering
         \includegraphics[scale=0.28]{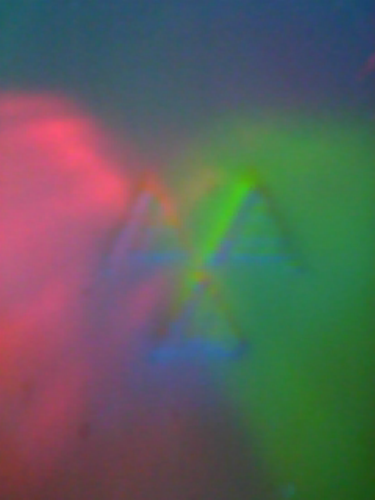}
         \caption{}
     \end{subfigure}
     \begin{subfigure}[b]{0.1\textwidth}
         \centering
         \includegraphics[scale=0.28]{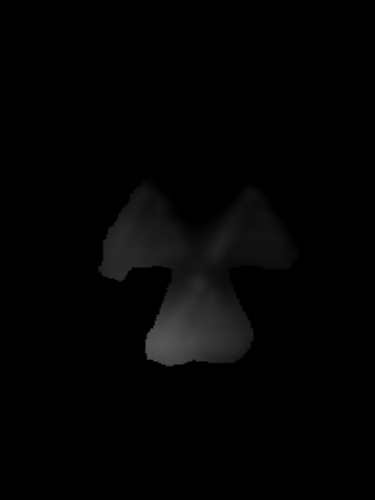}
         \caption{}
     \end{subfigure}

     \begin{subfigure}[b]{0.1\textwidth}
         \centering
         \includegraphics[scale=0.28]{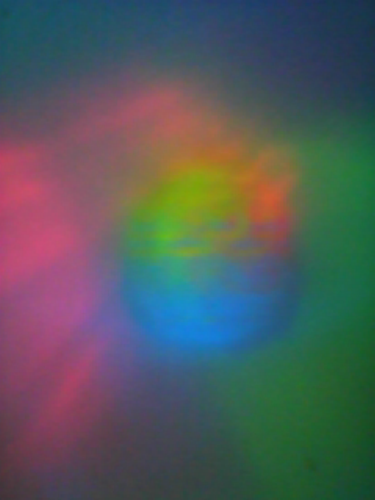}
         \caption{}
     \end{subfigure}
     \begin{subfigure}[b]{0.1\textwidth}
         \centering
         \includegraphics[scale=0.28]{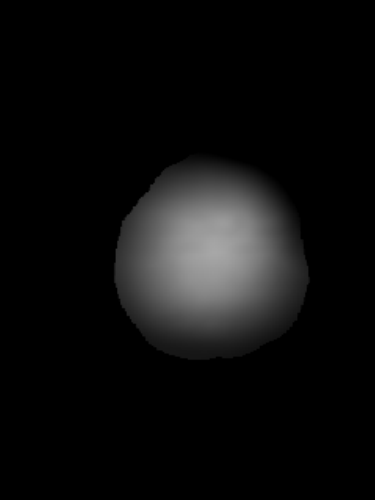}
         \caption{}
     \end{subfigure}
     \begin{subfigure}[b]{0.1\textwidth}
         \centering
         \includegraphics[scale=0.28]{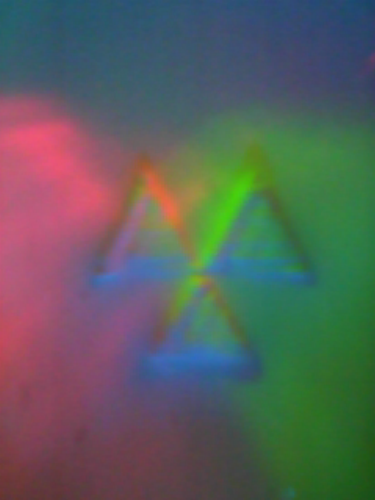}
         \caption{}
     \end{subfigure}
     \begin{subfigure}[b]{0.1\textwidth}
         \centering
         \includegraphics[scale=0.28]{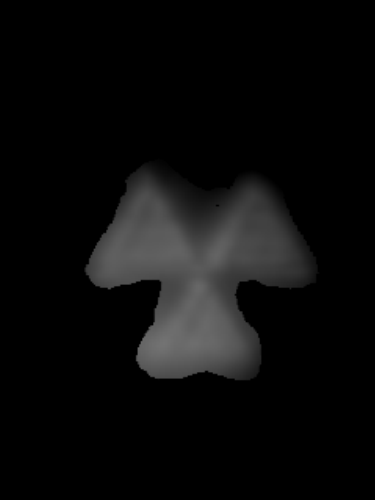}
         \caption{}
     \end{subfigure}

     \begin{subfigure}[b]{0.1\textwidth}
         \centering
         \includegraphics[scale=0.28]{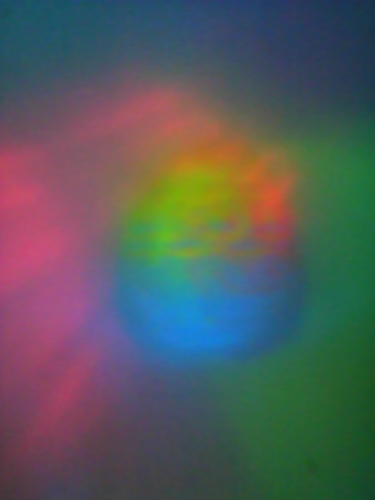}
         \caption{}
     \end{subfigure}
     \begin{subfigure}[b]{0.1\textwidth}
         \centering
         \includegraphics[scale=0.28]{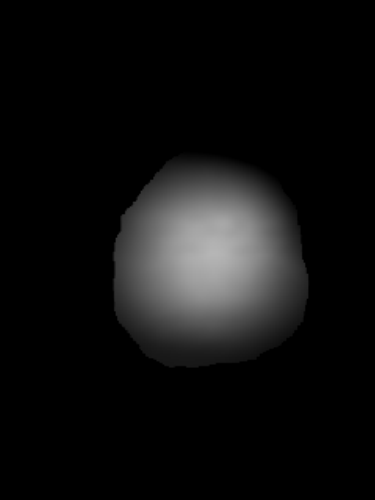}
         \caption{}
     \end{subfigure}
     \begin{subfigure}[b]{0.1\textwidth}
         \centering
         \includegraphics[scale=0.28]{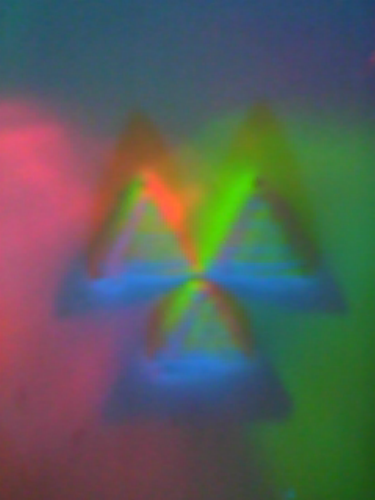}
         \caption{}
     \end{subfigure}
     \begin{subfigure}[b]{0.1\textwidth}
         \centering
         \includegraphics[scale=0.28]{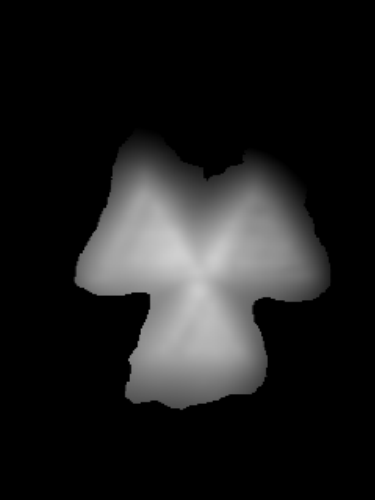}
         \caption{}
     \end{subfigure}

     \caption{(a,c,e,g,i,k) are RGB images. (b,d,f,h,j,l) are filtered depth images. The applied forces were: (a,b) 4.317N, (c,d) 3.485N, (e,f) 8.858N, (g,h) 6.535N, (i,j) 9.192N, and (k,l) 14.406N} 
     \label{fig:rgb_depth_examples}
\end{figure}

Some examples of the results of filtered depth images obtained from the RGB images are shown in Fig. \ref{fig:rgb_depth_examples}. In this figure, we also show how the two visuotactile representations change when the applied force is increased during the robotic grasping of two different objects. It is important to note that, although the calibration process is performed with a spherical object, the trained model can generalize and estimate depth maps for other objects with different geometries, sizes, and shapes as shown in Figure \ref{fig:rgb_depth_examples}.

\subsection{CNN Architectures for Normal Force Regression}

One of the objectives of this work is to determine which of the two types of visuotactile representation (RGB or depth) is the most effective in estimating normal grasping force.
This objective was achieved by implementing the following approaches: (i)  \textit{RGBmod}, which uses RGB data; (ii) \textit{D} and \textit{Dmod}, which use depth data; and (iii) \textit{RGBmod+D}, which uses both RGB and depth data. All the methods are based on the ResNet-18 architecture \cite{resnet} because it is fast and achieves a good performance in a wide range of tasks, including image regression.

\begin{figure}[htbp]
     \centering
         \centering
         \includegraphics[scale=0.34]{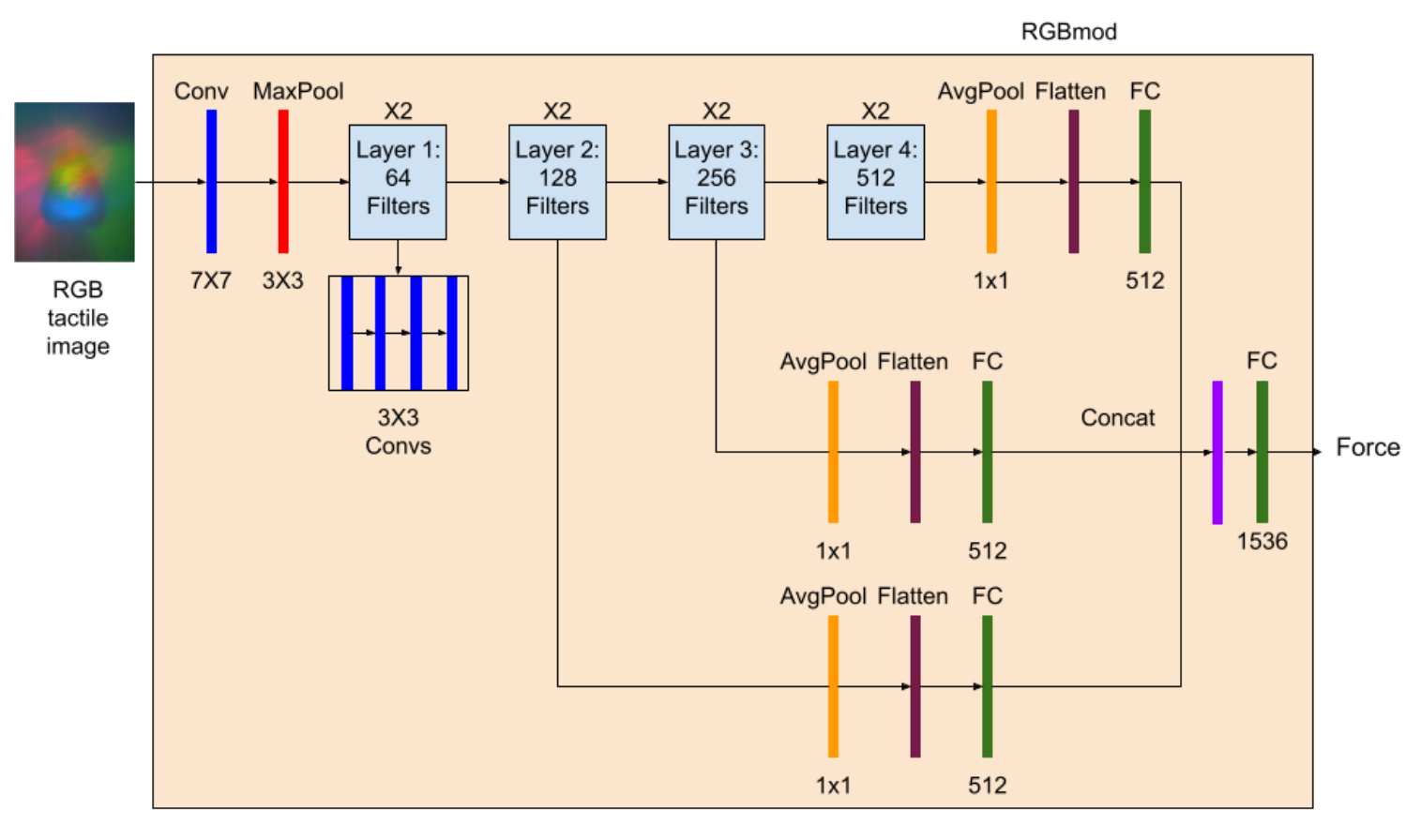}
     \caption{Our \textit{RGBmod} method estimates force values from a single RGB tactile image. The tactile features from the second, third, and last layer are concatenated to estimate the grasping force}
     \label{fig:modified_resnet18_arc}
\end{figure}

In the \textit{RGBmod} method, 
we modified the ResNet-18 architecture in order to estimate the force value from features extracted from different depth levels in the architecture (see Fig. \ref{fig:modified_resnet18_arc}), and not only from the last layer, as in the original version. This modification was made because tactile images are specifically composed of low-level features such as contours, textures, colors, etc. These types of features vanish as the number of layers in the architecture increases.

The \textit{D} method, whose input is the filtered depth image, is another of our modifications of the original ResNet architecture, in which the input RGB image is replaced by the filtered depth image generated from the color values, as explained above. Consequently, the dimensions of the first layer have been resized from 320x240x3 to 320x240x1. Besides, \textit{Dmod}, which is another method that estimates forces from the filtered depth image, differs from the \textit{D} method in that it uses the same technique to recover tactile features from early layers as in \textit{RGBmod}. 

\begin{figure}[htbp]
     \centering
         \centering
         \includegraphics[scale=0.5]{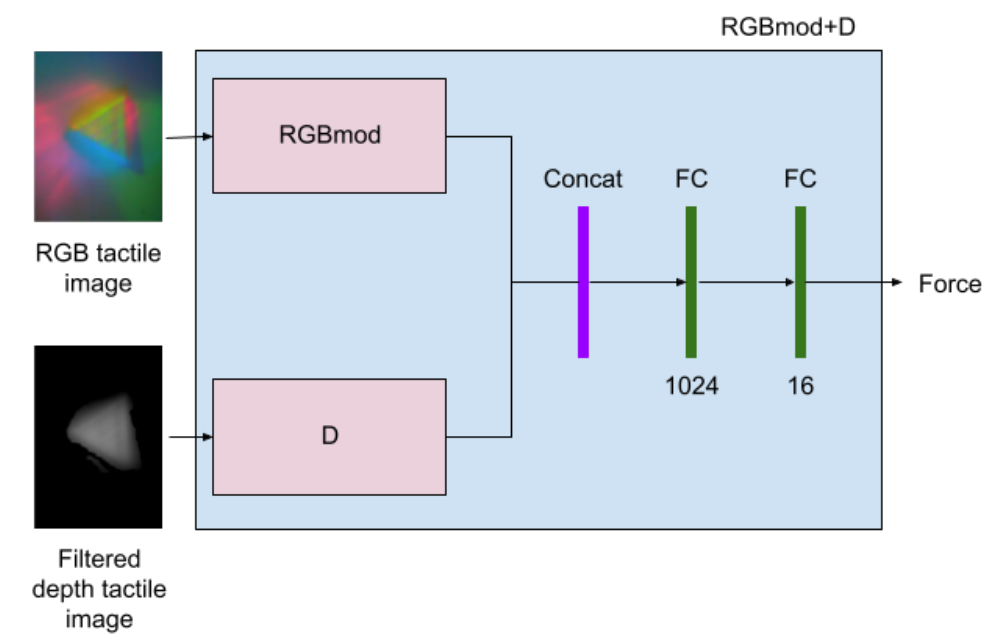}
     \caption{The \textit{RGBmod+D} method takes the RGB and the filtered depth tactile images as input, processes them separately using \textit{RGBmod} and \textit{D}, and concatenates the tactile features to estimate the force value} 
     \label{fig:rgbplusd}
\end{figure}

Finally, we have implemented \textit{RGBmod+D}, which combines RGB and depth tactile data by processing the RGB and depth images in separate branches so as to later concatenate the extracted features in order to regress the force value. A schematic of the \textit{RGBmod+D} architecture is shown in Fig. \ref{fig:rgbplusd}. 

\section{Experimental Setup}

\label{sec:experimental_setup}

\subsection{Robotic Setup with Tactile Sensors}
\begin{figure}[htbp]
     \centering
         \centering
         \includegraphics[scale=0.5]{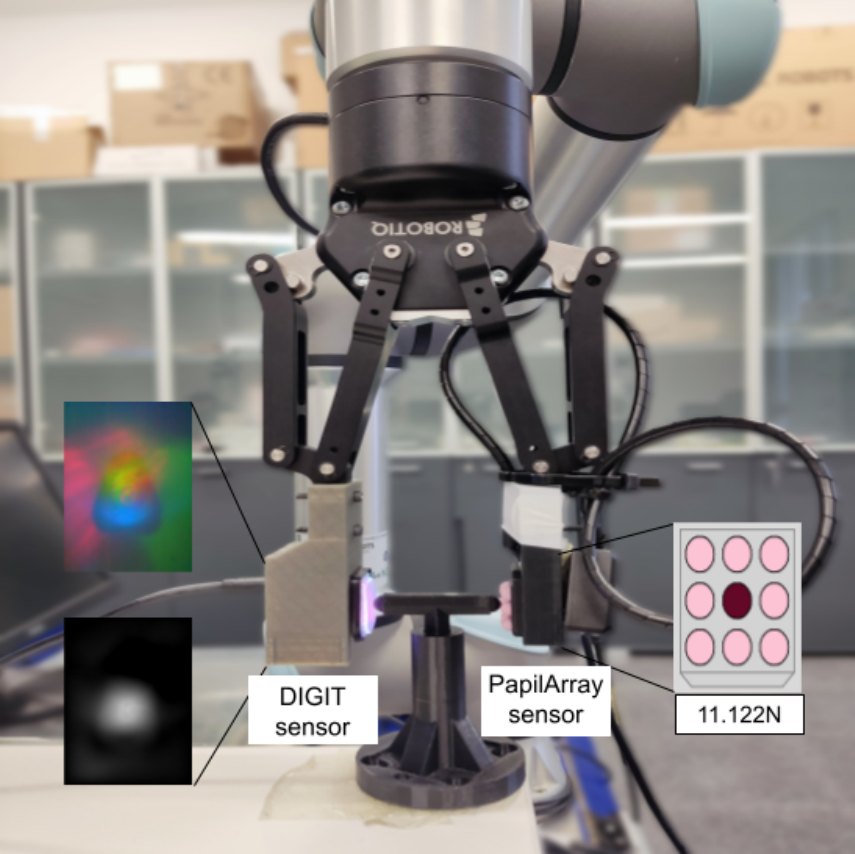}
     \caption{Example of our robotic setup during data collection, where the robot grasps one of the indenters in our dataset while the RGB, depth, and ground truth force data are recorded} 
     \label{fig:setup}
\end{figure}

Fig. \ref{fig:setup} shows our robotic setup, which consists of a UR5e robot and a ROBOTIQ-2F140 gripper. The gripper is equipped with a DIGIT sensor on one finger and a PapillArray sensor on the other finger in a parallel configuration. The object to be grasped is fixed to the table in order to avoid movement during the grasping process. In this work, we have defined our experimental setup in such a way that the ground truth measurements are generated by following three conditions. The first condition was that we designed the objects in our dataset (called indenters) to be symmetrical, to assume that the contact surfaces and forces would be identical for both sensors. The second condition was that the orientation of the robot was calibrated with respect to the indenter fixed on the table in order to achieve a perpendicular contact and a net force of zero. The last condition was that the distance between the surface of each sensor and the indenter was adjusted to ensure simultaneous contact of both sensors with the indenter.

In the literature cited above, it is common to use an F/T sensor to capture the ground truth forces in different ways, such as by placing the object on top of the F/T sensor and the tactile sensor on the robot end effector, as in \cite{2}, or by placing the tactile sensor on top of the F/T sensor and the object on the robot end effector, as in \cite{gelstereo}. An F/T force sensor does not directly measure the grasping force, but rather a single normal force at one point and in one direction. In contrast, in this work we have used a setup that is directly related to the grasping task, where the PapillArray sensor is placed on the opposite finger of the gripper. The use of this setup, therefore, allows us to obtain the grasping force applied to a larger contact area instead of a single point. 

\subsection{Data Acquisition and Dataset Generation}
\begin{figure}[htbp]
     \centering

        \begin{subfigure}[b]{
        0.48\textwidth}
        \centering
         \includegraphics[scale=0.7]{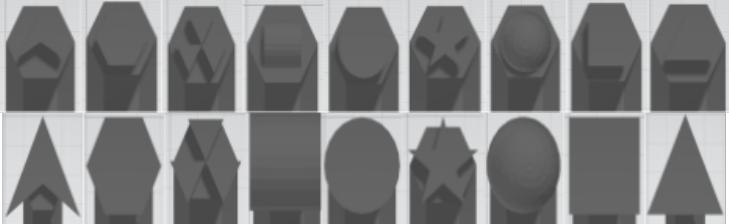}
         \caption{}
     \end{subfigure}
     
    \begin{subfigure}[b]{
    0.48\textwidth}
    \centering
         \includegraphics[scale=0.28]{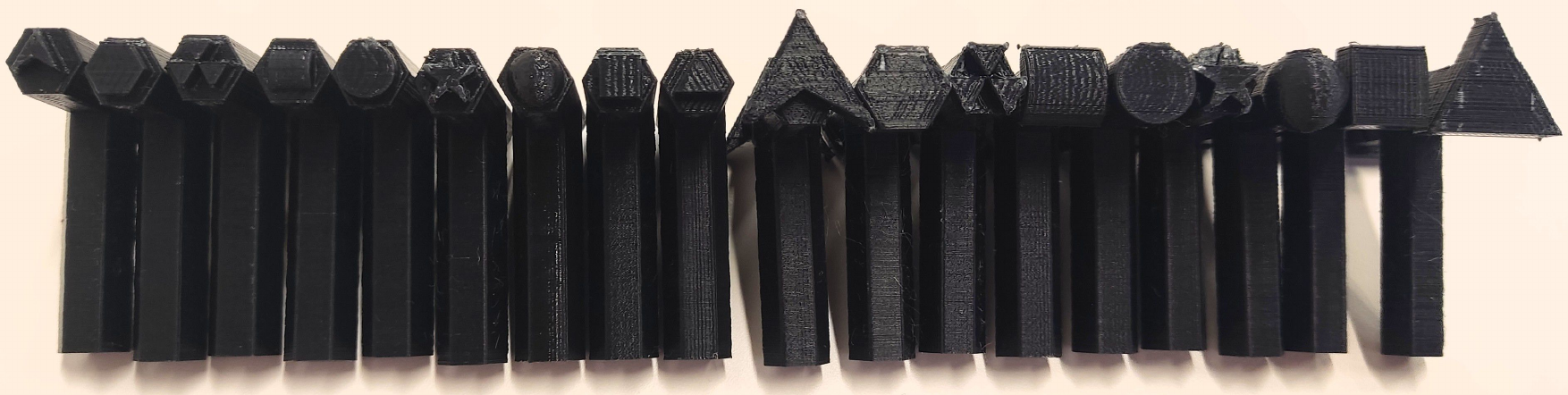}
         \caption{}
     \end{subfigure}

     \caption{(a) CAD models of the indenter geometries. (b) Indenters with different geometric shapes and scales. The first 9 indenters (from left to right) correspond to the first row in (a), and the last 9 indenters correspond to the second row in (a)} 
     \label{fig:indenters}
\end{figure}

We designed 18 objects, with different shapes (square, star, arrow, circle, etc.) and at two different scales, to generate symmetrical grasping points to collect our dataset. We called these objects as indenters, which are shown in Fig. \ref{fig:indenters}.
Although symmetrical contact surfaces are
required to obtain robust ground truth force data, our methods are not limited to measuring forces with symmetrical objects, as we will show in Section \ref{sec:non_symmetrical_objects}.

\begin{figure}[htbp]
     \centering
         \centering
         \includegraphics[scale=0.8]{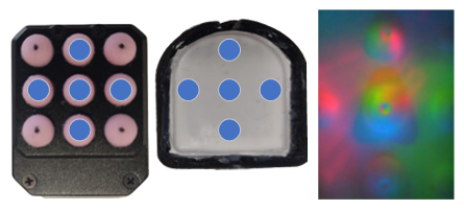}
     \caption{The blue dots indicate the positions where the indenters and the sensors made contact during data collection, avoiding partial contact or shear forces. The RGB tactile image on the right side is obtained by pressing both sensors together} 
     \label{fig:robot_trajectory}
\end{figure}

We collected our tactile data by controlling the robot along the surface of the sensors. During the trajectory, we grasped the indenters by the blue dots shown in Fig. \ref{fig:robot_trajectory}, while the sensor data (RGB and ground truth force data) were recorded. The positions of the blue dots are aligned with the PapillArray taxels to avoid partial contact or shear forces. In addition, the corner taxels of the PapillArray sensor were discarded to collect our dataset because they do not fit into the curved surface at the corners of the DIGIT sensors, as can be seen in Fig. \ref{fig:robot_trajectory}.

Our dataset consists of a total of 13,131 triplets of RGB images, depth images, and ground truth forces, varying the forces between 1N and 15N to avoid using negligible deformation values ($<$1N) or damaging the sensor ($>$15N).

In this work, we have applied a 3-fold cross-validation technique in order to randomly generate three different datasets (S1, S2, and S3) from the original dataset. Each dataset is divided into three sets (training, validation, and test) containing the samples of 14 randomly selected indenters for the training set, 2 random indenters for the validation set, and 2 random indenters for the test set.

\subsection{Metric Evaluation}
\label{sec:performance_evaluation}

In this work, our methods were evaluated using two different metrics, the Mean Absolute Error (MAE) and the Relative Error (RE). The MAE metric is commonly used to measure the performance of regression methods, but this metric calculates a mean absolute error value for all the samples without taking into account the magnitude of the ground truth force. For example, an error of 2N is considered the same whether the ground truth force is 3N or 15N. However, an error of 2N is significant if the ground truth force is 3N, but relatively small if it is 15N.  We consequently decided to use the RE metric as defined in Eq. \ref{eq:relative_error}, obtaining a force error relative to the applied force magnitude,

\begin{equation}
    RE = \frac{|f_{pred}-f_{gt}|}{f_{gt}}
    \label{eq:relative_error}
\end{equation}

where $f_{pred}$ is the predicted force regression of only one sample, and $f_{gt}$ is the corresponding ground truth force. 

Moreover, the proposed methods are trained with the following hyperparameters: a batch size of 64, a learning rate of 4e-5, the Adam optimizer, 25 training epochs, weights from scratch, and no data augmentation is applied. These hyperparameters were chosen following a random search procedure, performed in order to optimize the training results. We used an NVIDIA GeForce RTX 3060 with 12GB of RAM for both the training and the test phases.

\section{Experimental Results and Analysis}

\label{sec:experimental_results_analysis}
\subsection{Evaluation of the performance of the proposed methods}

The test phase results obtained after the training and validation phases are shown in Table \ref{tab:experiment1}.


\begin{table*}[htpb]
 \centering
 \caption{Results in terms of the mean and standard deviation of the RE for our approaches using the test set of each dataset. Note that the RE is dimensionless and the MAE is measured in N}
 \label{tab:experiment1}
\begin{tabular}{|c|cc|cc|cc|}
    \hline
    \textbf{Test sets} & \multicolumn{2}{c|}{\textbf{S1}}                     & \multicolumn{2}{c|}{\textbf{S2}}                     & \multicolumn{2}{c|}{\textbf{S3}}                     \\ \hline
    \backslashbox{\textbf{Methods}}{\textbf{Metrics}}         & \multicolumn{1}{c|}{\textbf{RE}}    & \textbf{MAE}   & \multicolumn{1}{c|}{\textbf{RE}}    & \textbf{MAE}   & \multicolumn{1}{c|}{\textbf{RE}}    & \textbf{MAE}   \\ \hline
    \textbf{RGBmod}   & \multicolumn{1}{c|}{0.173 $\pm$ 0.306} & 0.931 $\pm$ 0.798 & \multicolumn{1}{c|}{0.206 $\pm$ 0.392} & 1.035 $\pm$ 1.074 & \multicolumn{1}{c|}{0.222 $\pm$ 0.447} & 1.004 $\pm$ 0.847 \\ \hline
    \textbf{D}        & \multicolumn{1}{c|}{0.911 $\pm$ 1.631} & 3.010 $\pm$ 2.245& \multicolumn{1}{c|}{0.949 $\pm$ 1.689} & 3.107 $\pm$ 2.447& \multicolumn{1}{c|}{0.920 $\pm$ 1.626} & 3.028 $\pm$ 2.347\\ \hline
    \textbf{RGBmod+D} & \multicolumn{1}{c|}{0.186 $\pm$ 0.279} & 1.094 $\pm$ 0.863& \multicolumn{1}{c|}{0.205 $\pm$ 0.389} & 1.018 $\pm$ 1.023& \multicolumn{1}{c|}{0.213 $\pm$ 0.433} & 0.972 $\pm$ 0.868\\ \hline
    \textbf{Dmod}     & \multicolumn{1}{c|}{0.867 $\pm$ 1.501} & 3.248 $\pm$ 2.343 & \multicolumn{1}{c|}{0.919 $\pm$ 1.623} & 3.135 $\pm$ 2.373 & \multicolumn{1}{c|}{0.903 $\pm$ 1.603} & 3.221 $\pm$ 2.458 \\ \hline
\end{tabular}
\end{table*}

First, note that the methods based only on depth information obtained the highest RE and MAE values compared to the methods based on RGB data. Furthermore, \textit{D} and \textit{Dmod} obtained similar errors for the different datasets and metrics. The limitations of depth-based methods are discussed in Section \ref{sec:limitations_depth}.

In contrast, \textit{RGBmod}, the approach using only RGB data, obtained good results on the three datasets, with an RE of 0.173 and an MAE of 0.931N in the best case.
Note that \textit{RGBmod+D}, which combines RGB and depth data, is not significantly better than \textit{RGBmod}. Both approaches obtained errors of the same order for all the sets and metrics. However, \textit{RGBmod} runs 18.71ms faster than \textit{RGBmod+D.}
These results lead us to believe that the depth information does not help to learn a better representation and mapping with which to estimate the grasping forces.  
Therefore, we believe that the slight improvement of \textit{RGBmod+D} does not justify the complexity of generating and filtering the depth tactile images in real time.

\begin{figure}[htbp]
     \centering
         \centering
         \includegraphics[scale=0.4]{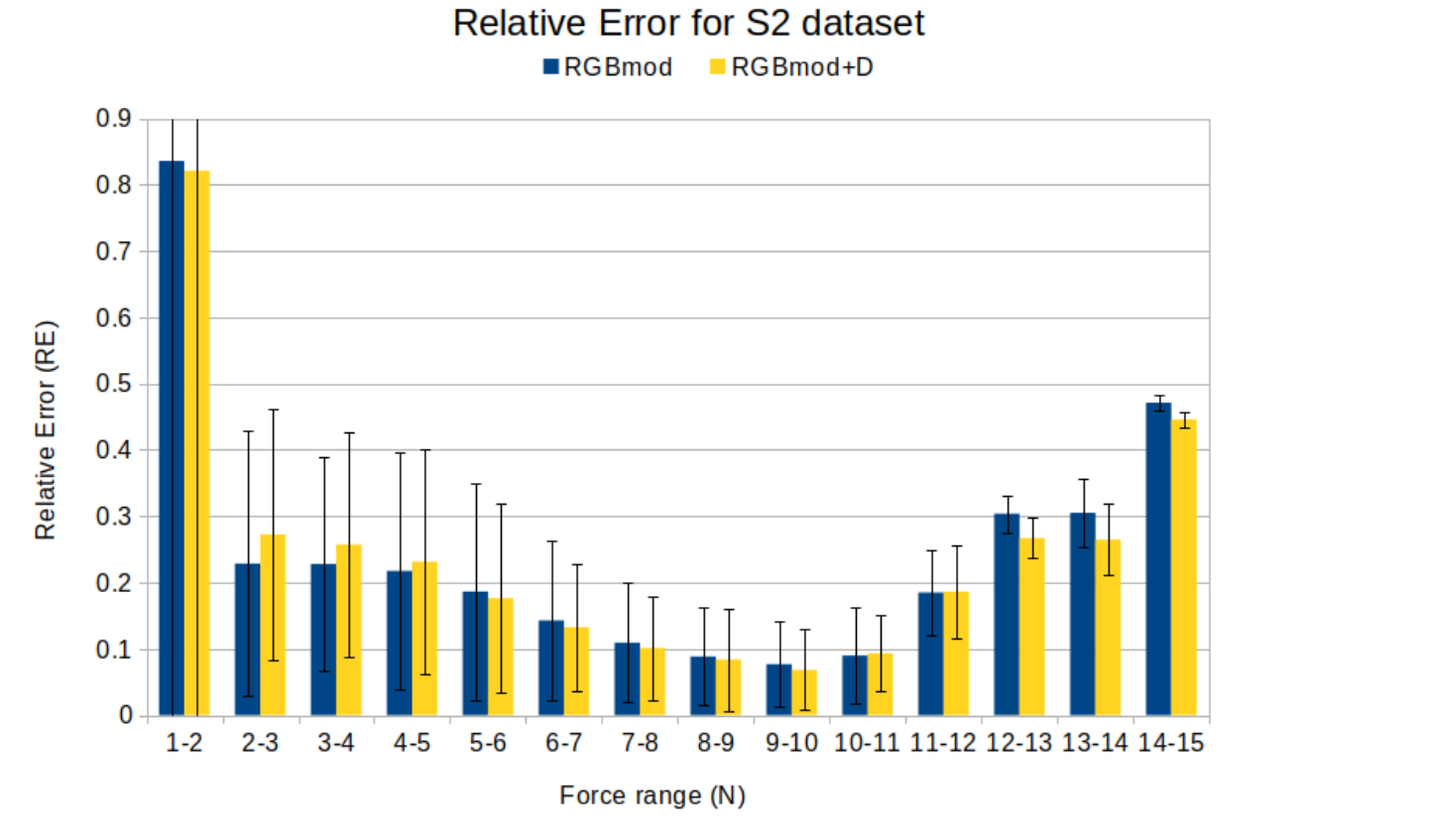}
     \caption{RE trend by force range for \textit{RGBmod} and \textit{RGBmod+D} with the S2 test set. The trend is also similar for the S1 and S3 test sets}
     \label{fig:comparacion_rangos_fuerza}
\end{figure}

In Table \ref{tab:experiment1}, we obtained the results for the different approaches using the full range of forces from 1 to 15N. In the following experiment, we aim to evaluate the proposed approaches with different force ranges to observe how the error changes with the applied force. 
This was done by testing our proposals (\textit{RGBmod} and \textit{RGBmod+D}) with the S2 test set, and calculating the RE metric in force intervals of 1N ([1-2)N, [2-3)N, etc.) as shown in Fig. \ref{fig:comparacion_rangos_fuerza}.

Several conclusions can be drawn from this experiment. First, we confirmed that \textit{RGBmod+D} does not improve the performance when compared to \textit{RGBmod}. 
Second, the RE values are quite high for forces between 1 and 2N because some properties of the gel, such as the hardness, make the deformation undetectable to the camera in this force range. Note that the RE values and the standard deviation decrease as the force increases. However, the DIGIT sensors tend to saturate at high forces, in this case above 15N. The DIGIT gel can deform with the applied force until a certain force value is reached. As the force increases, the gel will no longer deform, meaning that the tactile features captured by the sensor will remain the same even if a higher force is applied.

\subsection{Limitations of depth data for force regression task}
\label{sec:limitations}

Our experiments from Table \ref{tab:experiment1} and Fig. \ref{fig:comparacion_rangos_fuerza} show that depth data does not improve the performance in the force estimation task. This can be due to the following reasons.

First, it is important to consider that the surface of the DIGIT sensors is not completely flat, but curved. This curvature means that the deformation is not identical on every part of the surface of the sensor. In addition to the curvature of the surface, the geometric features of the object also cause different deformations. For example, a large planar indenter will generate less deformation
than a small spherical indenter, because the applied force is more evenly distributed over the planar surface than over the spherical surface.

\label{sec:limitations_depth}
\begin{figure}[htbp]
     \centering
         \centering
         \includegraphics[scale=0.6]{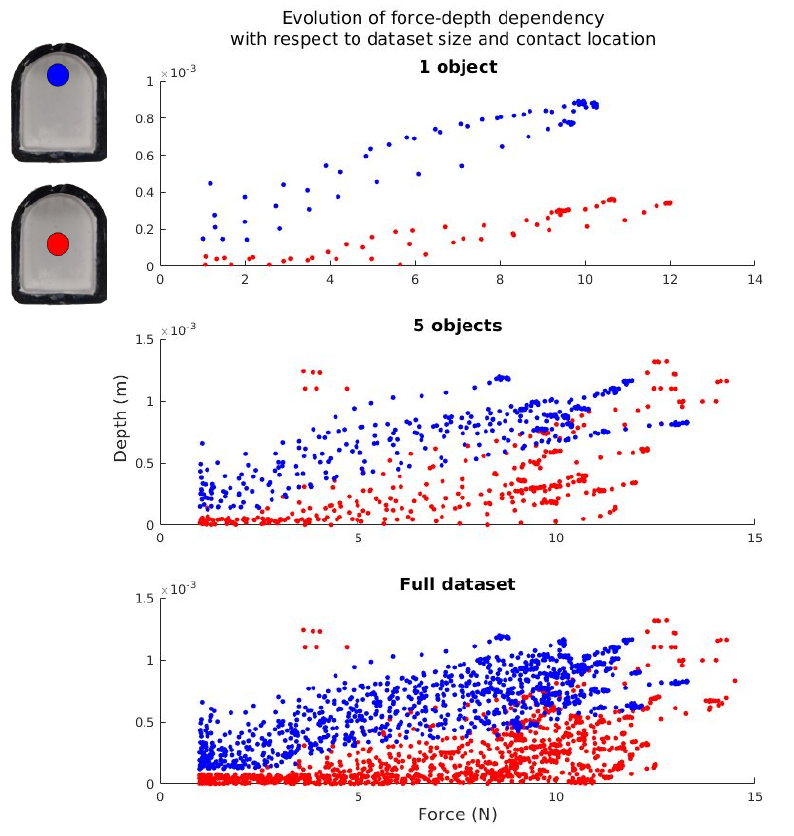}
     \caption{Evolution of force-depth mathematical relationship with different dataset sizes and contact areas (center and top) for the same DIGIT sensor unit. Depth is obtained as the maximum deformation value of the depth image for each applied force}
     \label{fig:limitations_depth}
\end{figure}

The purpose of Fig. \ref{fig:limitations_depth} is to show the trend of the force-depth mathematical relationship with different numbers of objects (indenters) and two contact surfaces of a DIGIT sensor. 
The trend is approximately linear when only one indenter and one contact area are considered. However, as we increase the number of indenters in the dataset, this relationship becomes very noisy. Also note that it is more linear for the top contact surface compared to the center of the surface. This is because the force is better distributed in planar surfaces. The surface is flatter in the middle than at the top, so the planar indenters cause less deformation for the same applied force. In summary, the force-depth relationship tends to be linear for one indenter and one contact surface. When considering different objects, contact surfaces, and even sensors, it is very complex to model.

\subsection{Experimentation with everyday objects}

In this section, our goal is to test our best method (\textit{RGBmod}) by grasping 10 unseen everyday objects with different contact forces and locations. We performed three grasps per object and DIGIT sensor, for a total of 60 grasps and 8,554 force samples. With these considerations and input forces from 5 up to 11N, where the best performance is obtained as shown in Fig. \ref{fig:comparacion_rangos_fuerza}, we calculated the RE metric to evaluate our method, and obtained an average value of 0.125 $\pm$ 0.153.

\begin{figure*}[htbp]
     \centering

     \begin{subfigure}[b]{0.23\textwidth}
         \centering
         \includegraphics[scale=0.24]{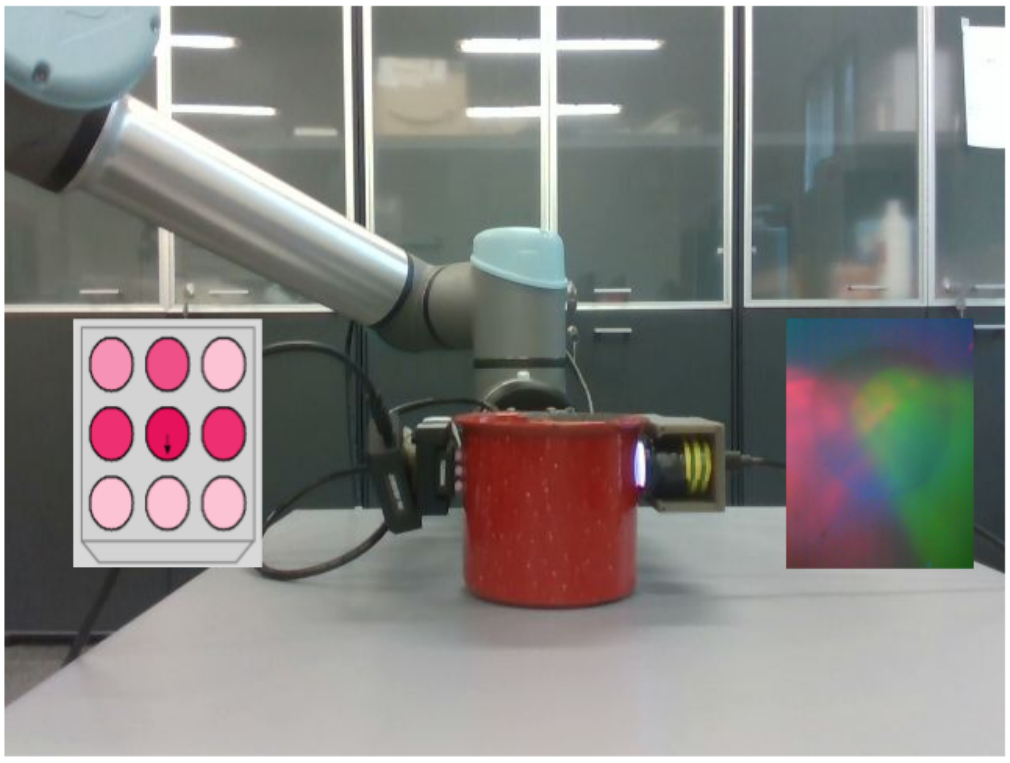}
     \end{subfigure}
    \begin{subfigure}[b]{0.23\textwidth}
         \centering
         \includegraphics[scale=0.28]{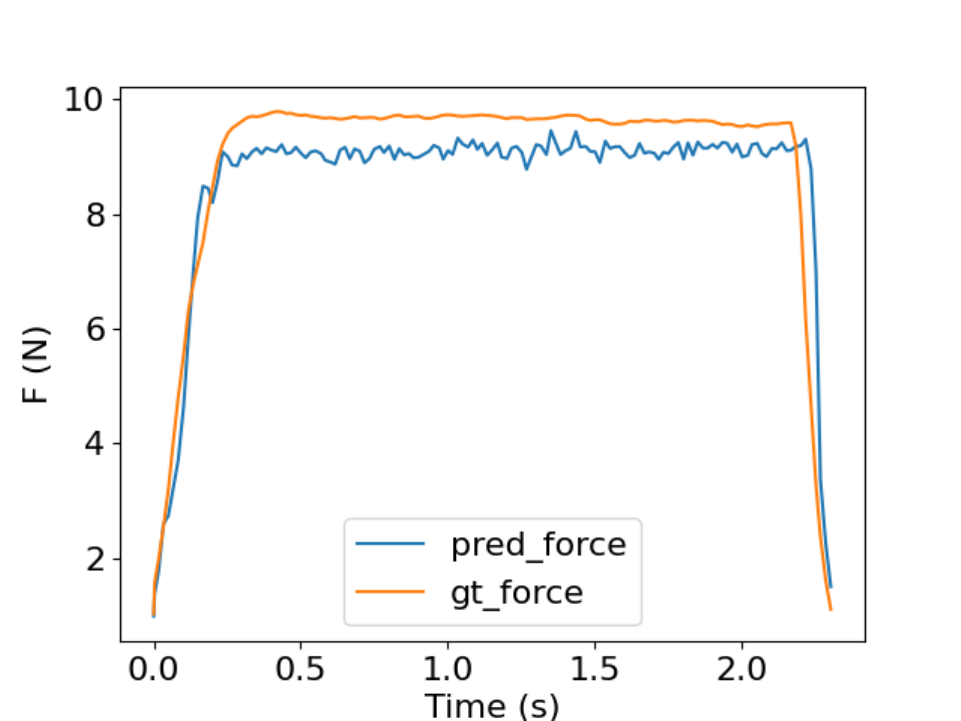}
     \end{subfigure}
      \begin{subfigure}[b]{0.23\textwidth}
         \centering
         \includegraphics[scale=0.24]{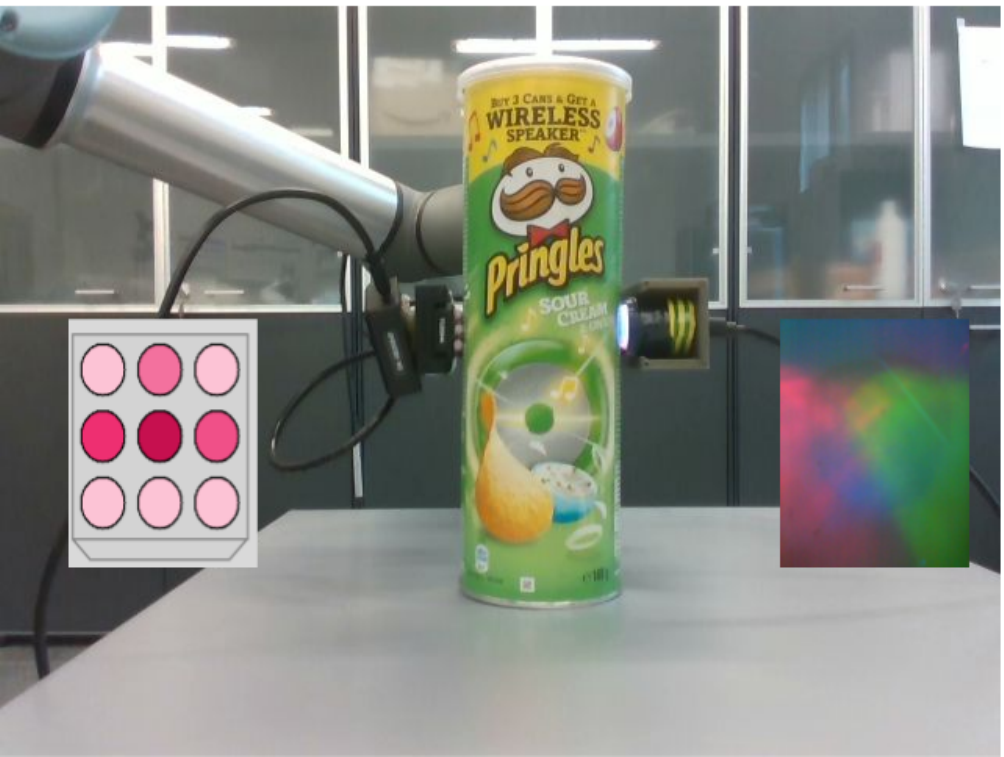}
     \end{subfigure}
    \begin{subfigure}[b]{0.23\textwidth}
         \centering
         \includegraphics[scale=0.28]{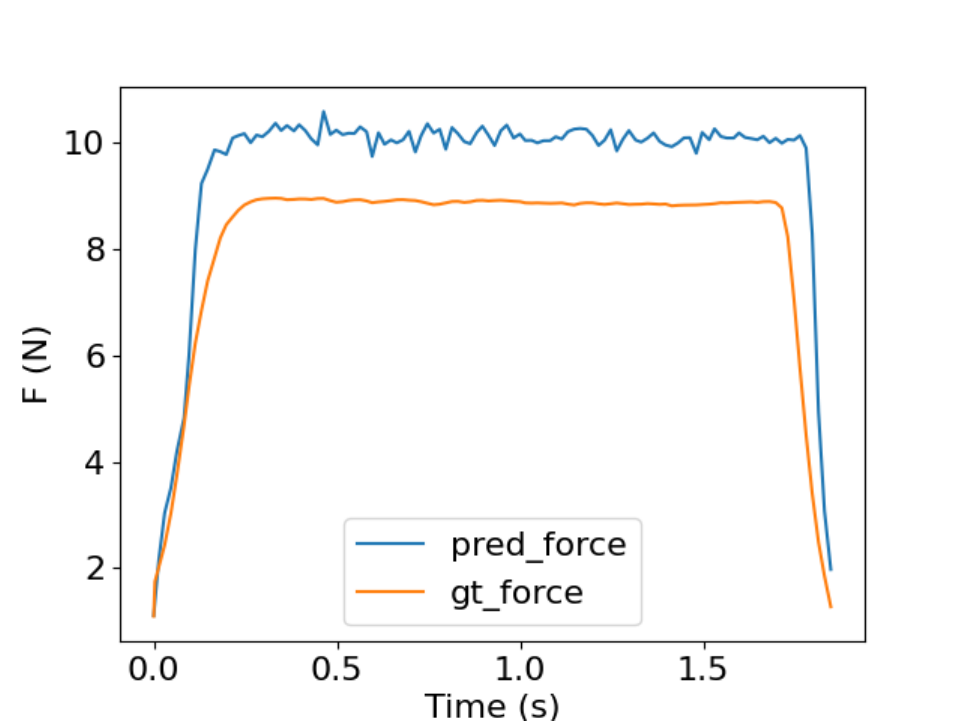}
     \end{subfigure}

     \begin{subfigure}[b]{0.23\textwidth}
         \centering
         \includegraphics[scale=0.24]{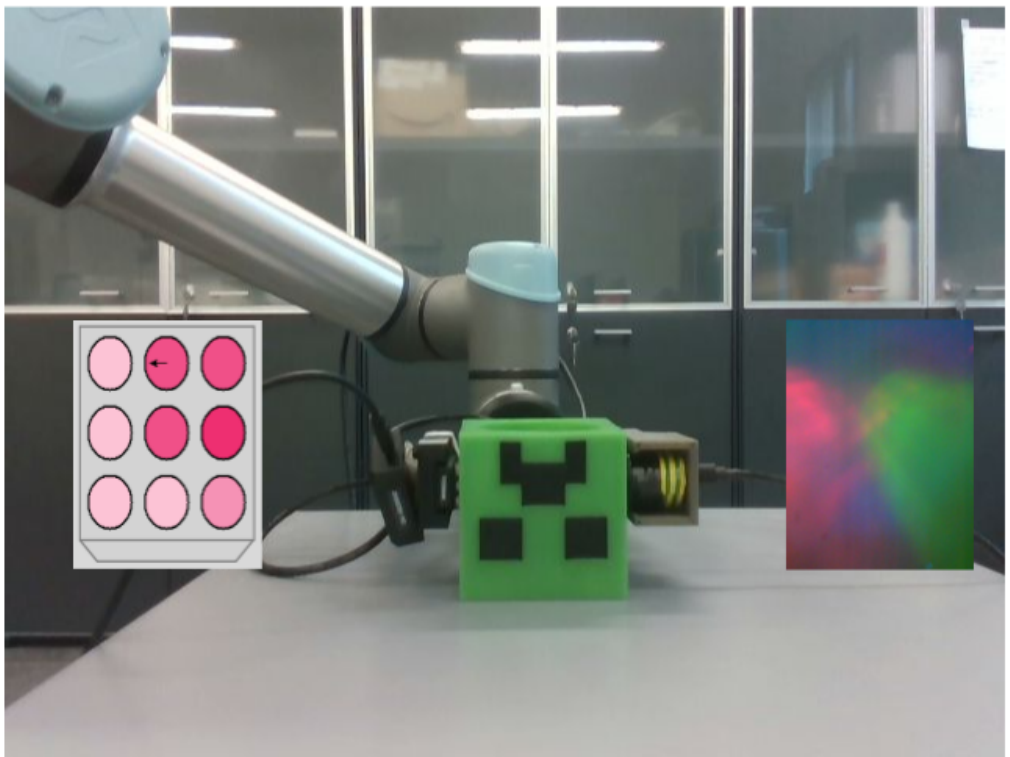}
     \end{subfigure}
    \begin{subfigure}[b]{0.23\textwidth}
         \centering
         \includegraphics[scale=0.28]{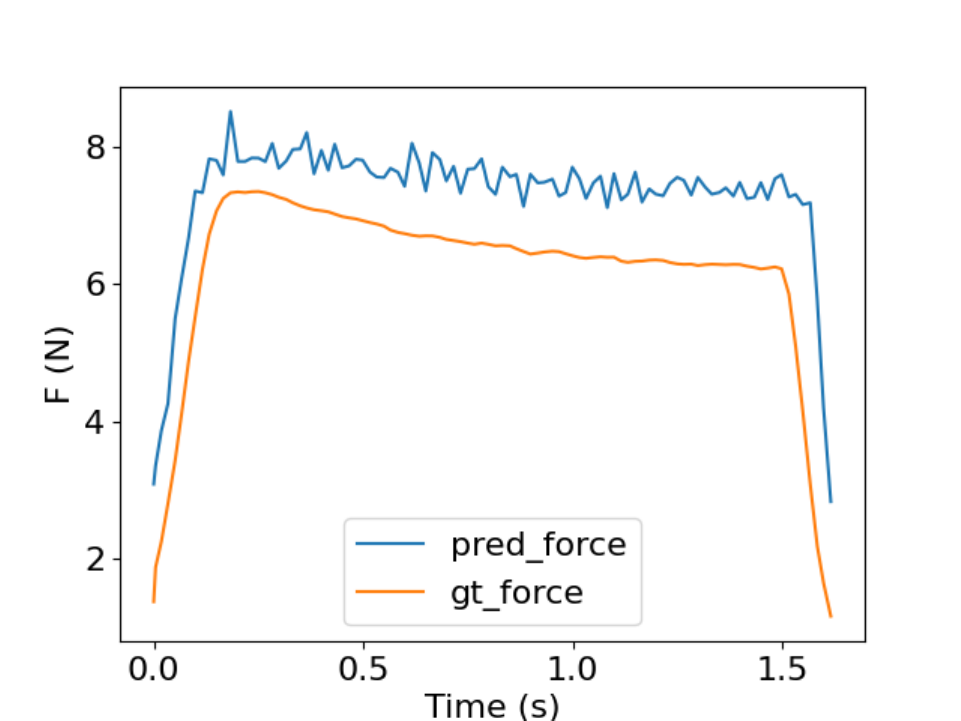}
     \end{subfigure}
     \begin{subfigure}[b]{0.23\textwidth}
         \centering
         \includegraphics[scale=0.24]{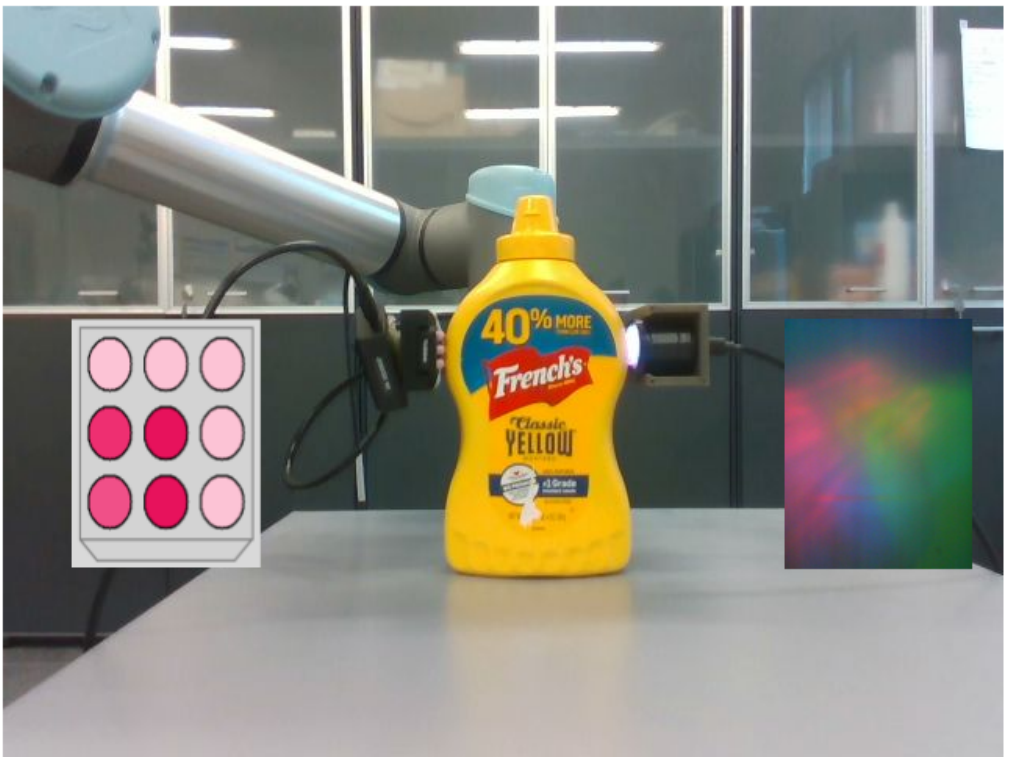}
     \end{subfigure}
    \begin{subfigure}[b]{0.23\textwidth}
         \centering
         \includegraphics[scale=0.28]{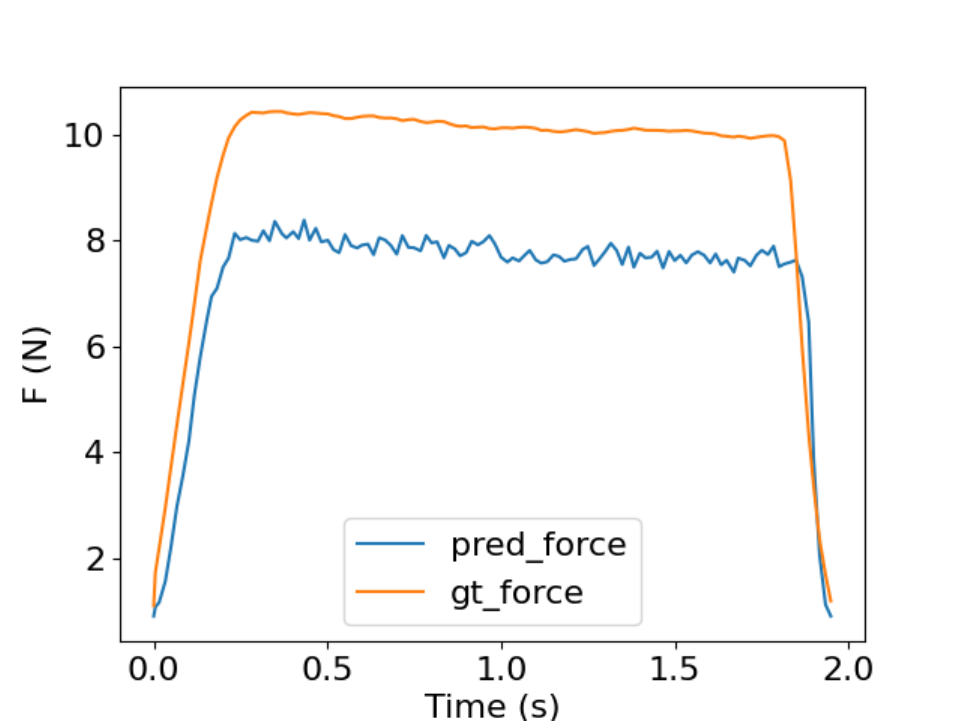}
     \end{subfigure}

     \begin{subfigure}[b]{0.23\textwidth}
         \centering
         \includegraphics[scale=0.24]{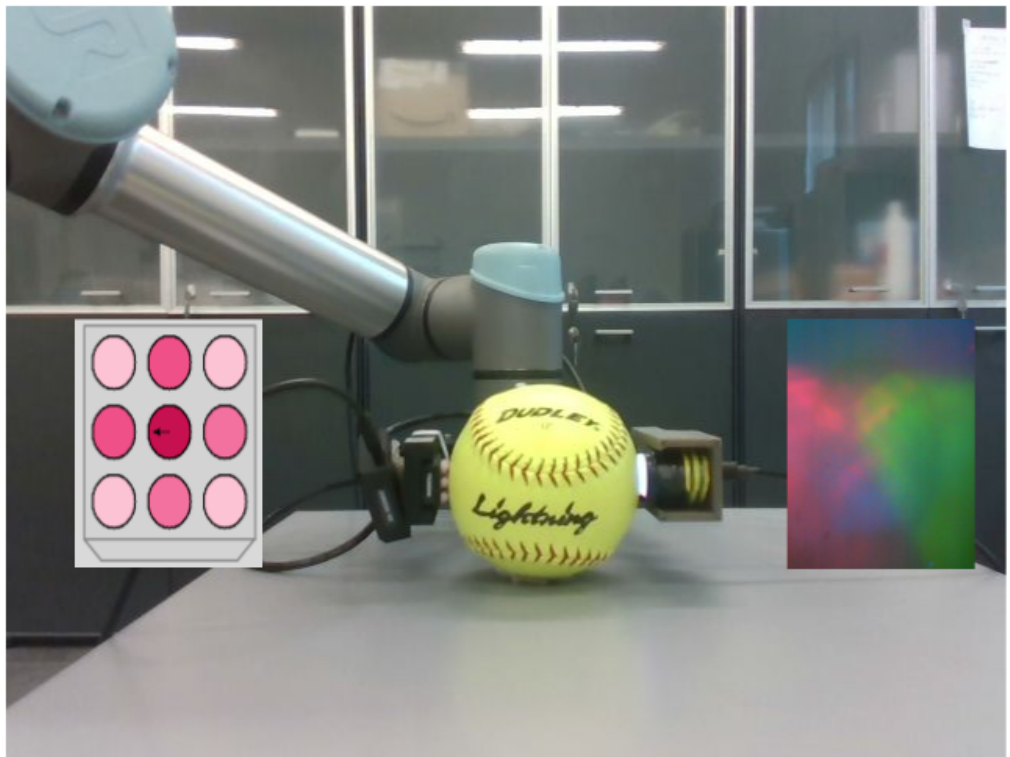}
     \end{subfigure}
    \begin{subfigure}[b]{0.23\textwidth}
         \centering
         \includegraphics[scale=0.28]{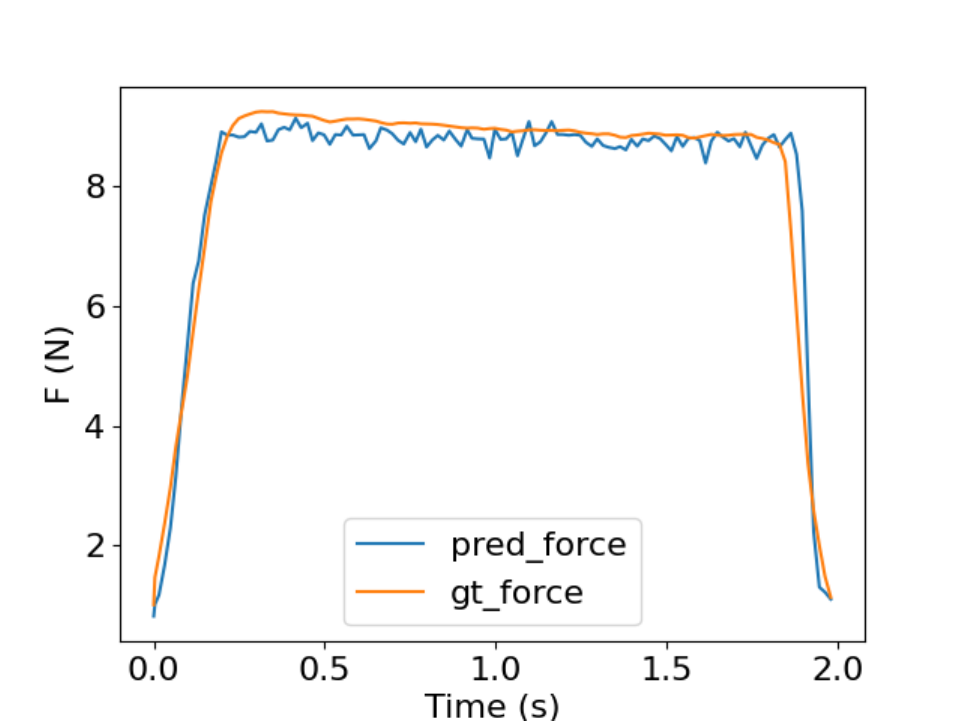}
     \end{subfigure}
     \begin{subfigure}[b]{0.23\textwidth}
         \centering
         \includegraphics[scale=0.24]{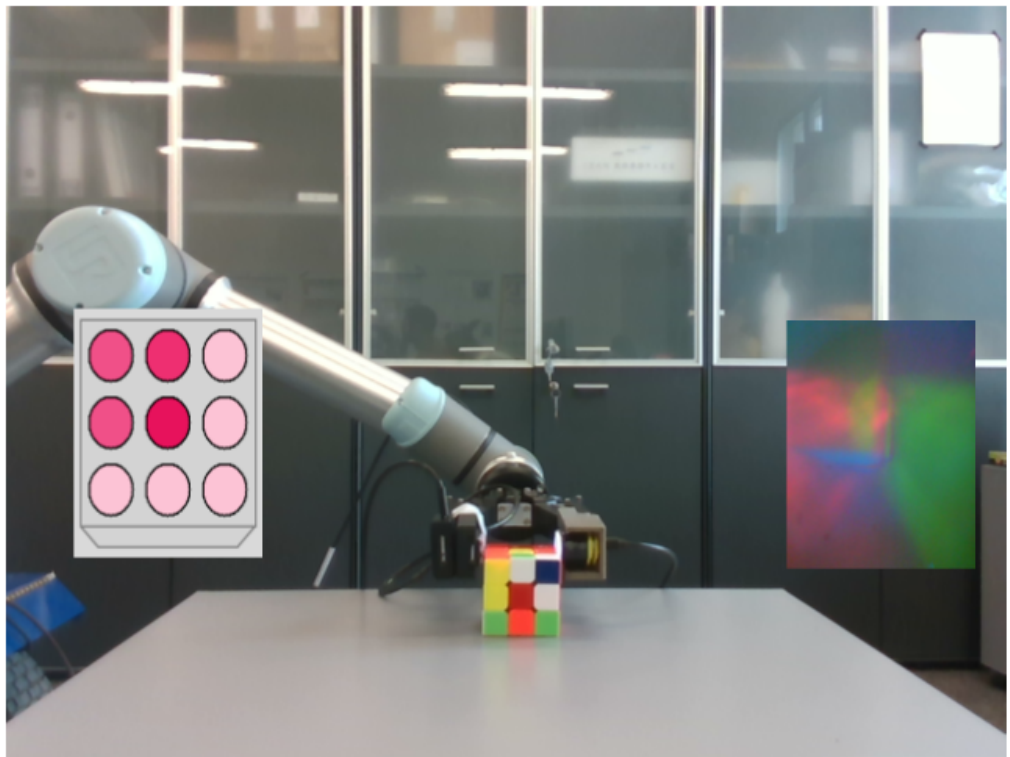}
     \end{subfigure}
    \begin{subfigure}[b]{0.23\textwidth}
         \centering
         \includegraphics[scale=0.28]{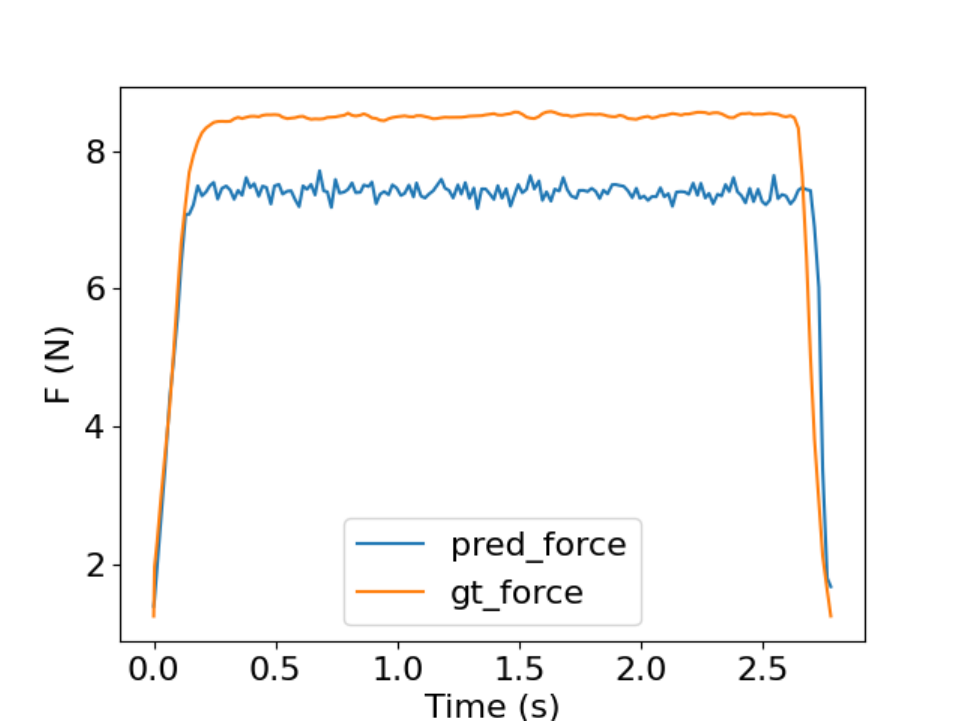}
     \end{subfigure}
     
     \begin{subfigure}[b]{0.23\textwidth}
         \centering
         \includegraphics[scale=0.24]{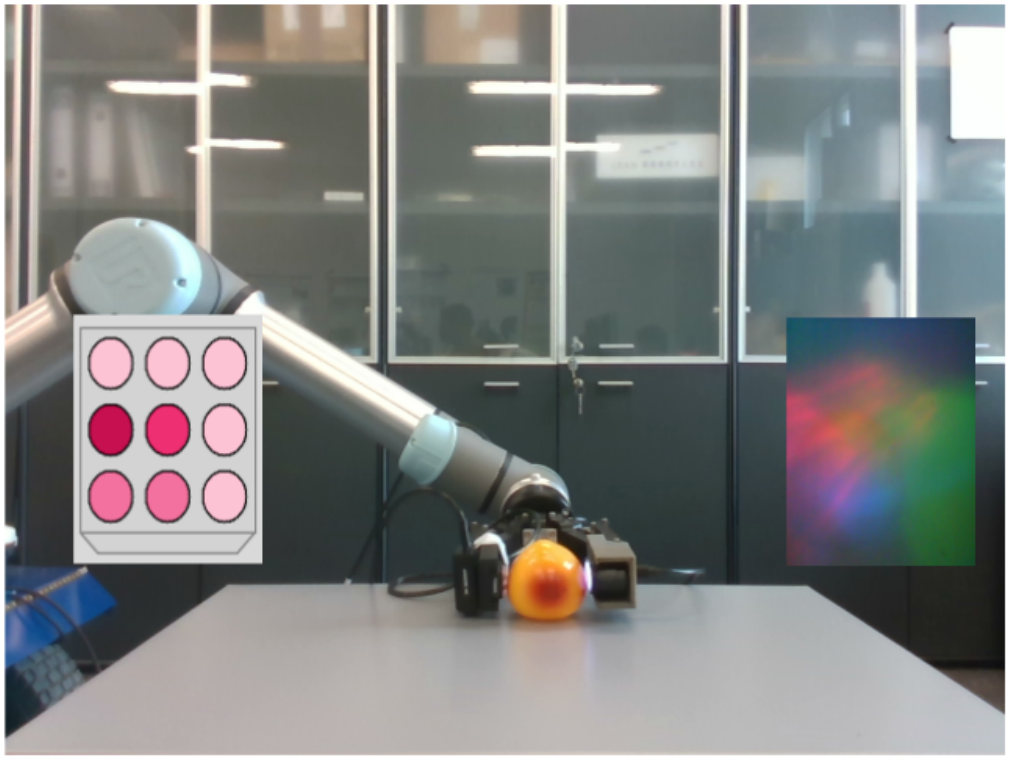}
     \end{subfigure}
    \begin{subfigure}[b]{0.23\textwidth}
         \centering
         \includegraphics[scale=0.28]{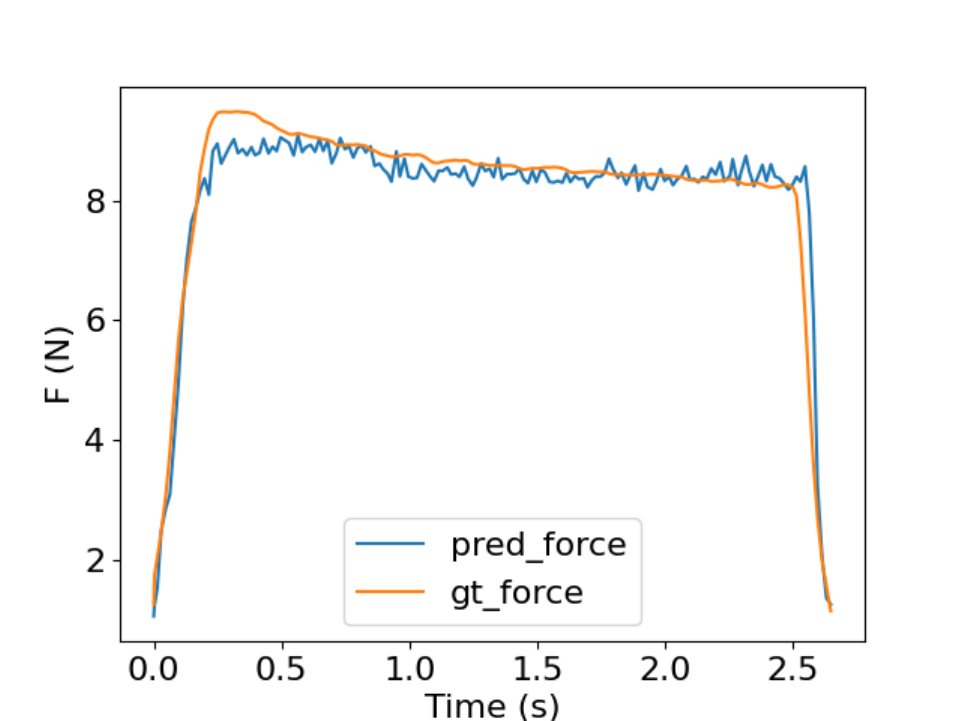}
     \end{subfigure}
      \begin{subfigure}[b]{0.23\textwidth}
         \centering
         \includegraphics[scale=0.24]{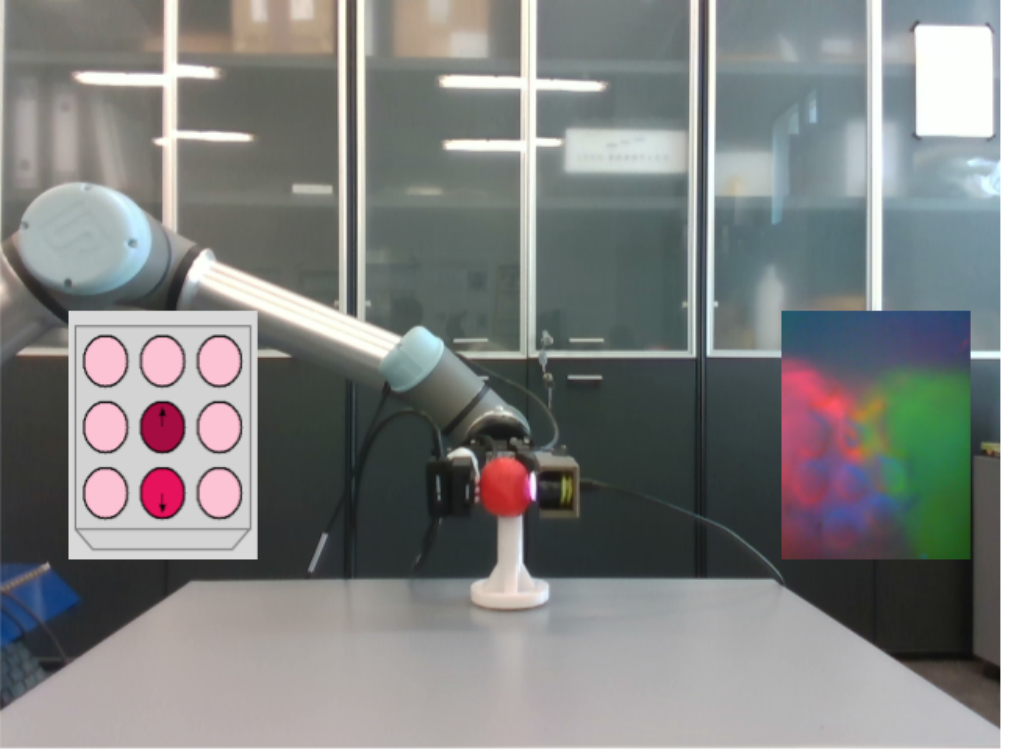}
     \end{subfigure}
    \begin{subfigure}[b]{0.23\textwidth}
         \centering
         \includegraphics[scale=0.28]{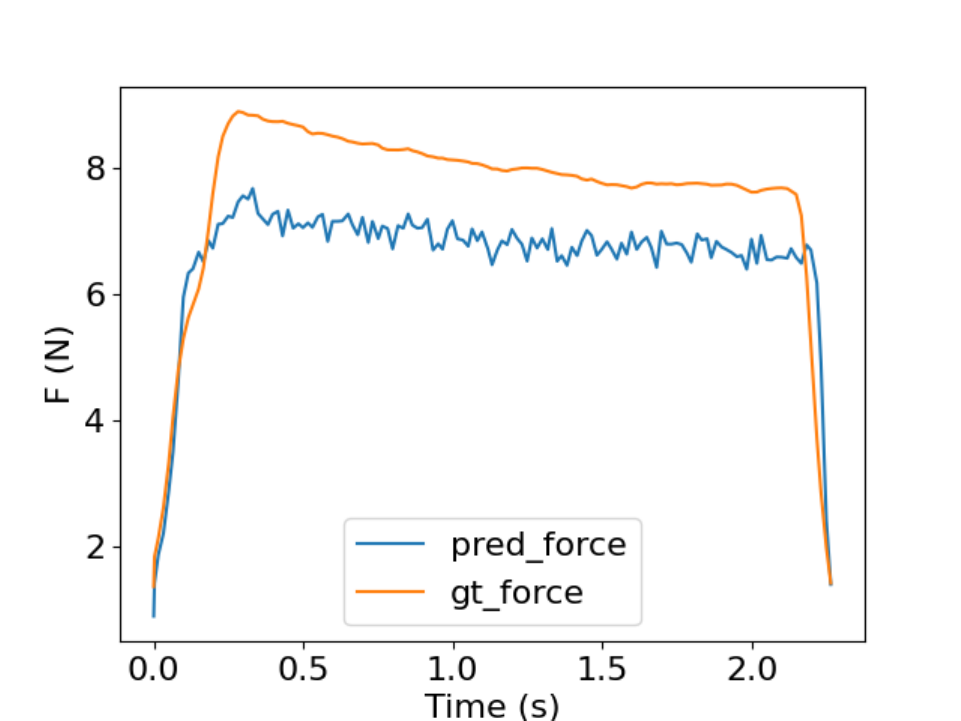}
     \end{subfigure}
     
     \begin{subfigure}[b]{0.23\textwidth}
         \centering
         \includegraphics[scale=0.24]{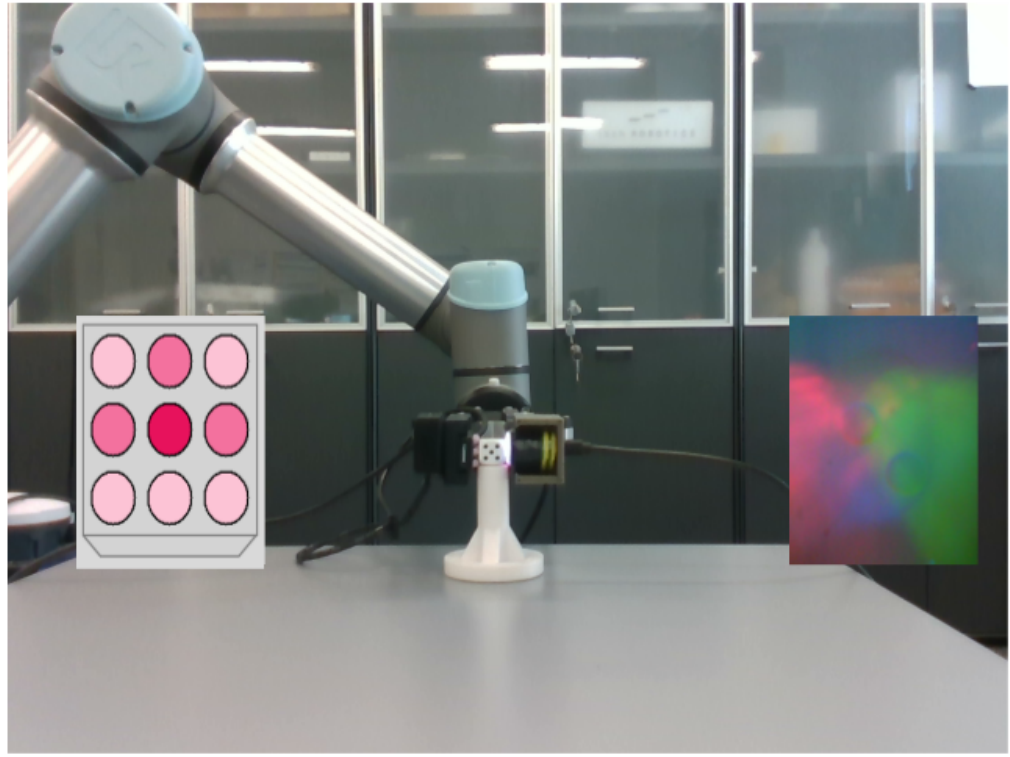}
     \end{subfigure}
    \begin{subfigure}[b]{0.23\textwidth}
         \centering
         \includegraphics[scale=0.28]{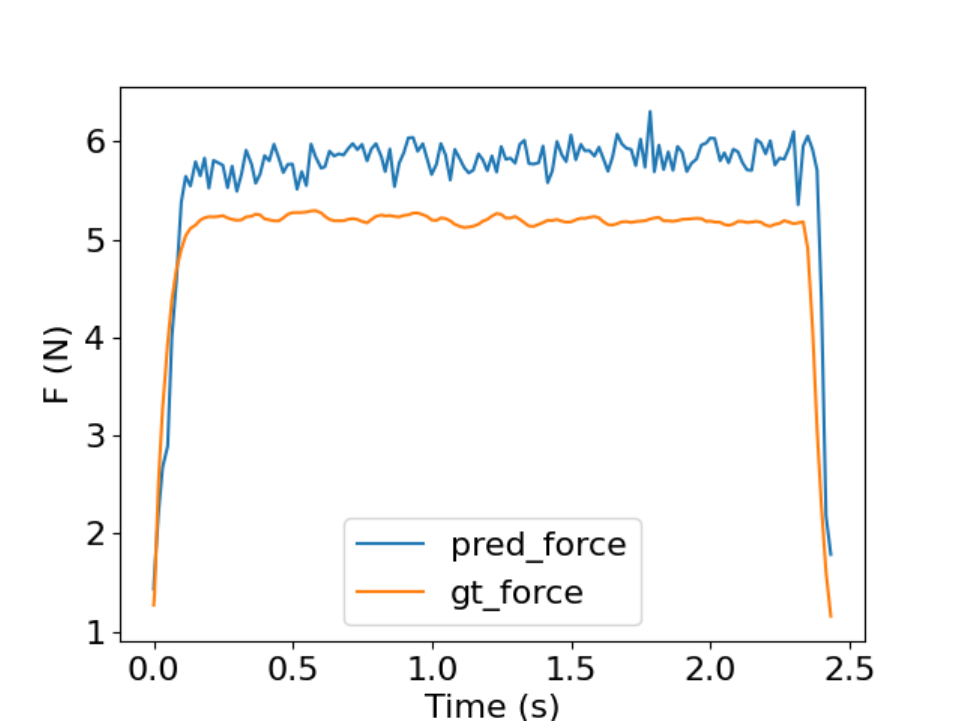}
     \end{subfigure}
     \begin{subfigure}[b]{0.23\textwidth}
         \centering
         \includegraphics[scale=0.24]{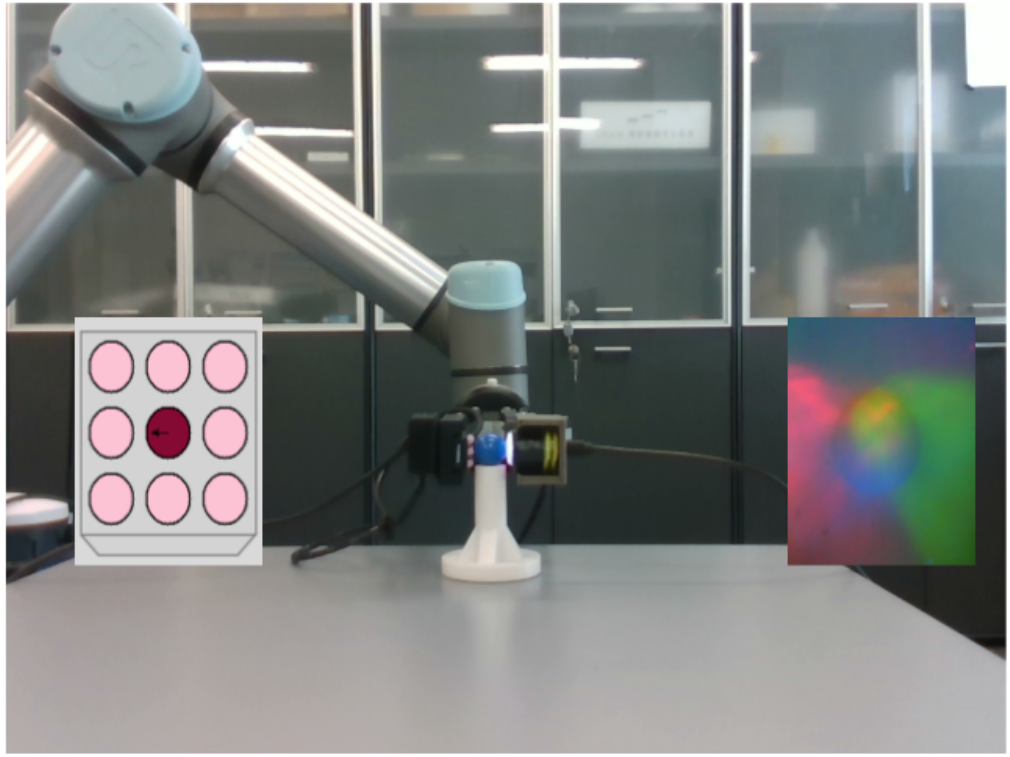}
     \end{subfigure}
    \begin{subfigure}[b]{0.23\textwidth}
         \centering
         \includegraphics[scale=0.28]{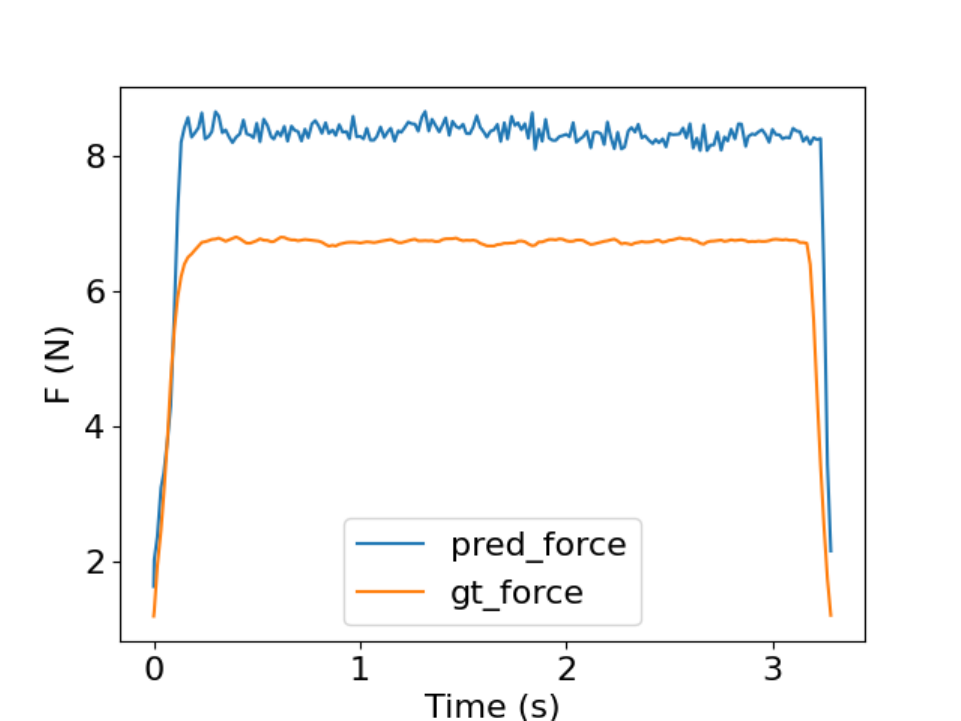}
     \end{subfigure}
   
      \caption{Examples of the estimated normal forces with \textit{RGBmod} when several unseen everyday objects are grasped with different applied forces, contact locations, and DIGIT sensors }
     \label{fig:everyday_objects_examples}
\end{figure*}

In more detail, Table \ref{tab:experiment2} shows the RE values obtained for each of the objects shown in Fig. \ref{fig:everyday_objects_examples}. The lowest RE (0.068 $\pm$ 0.103) was obtained with the Soccer ball while the highest RE (0.183 $\pm$ 0.174) was observed with the Marble. Note that the objects used in this experiment are made of different materials, resulting in different properties, such as rigidity or size. Therefore, we also analyzed the results from Table \ref{tab:experiment2} taking into account these properties.

\begin{table*}[htpb]
\centering
\caption{Results in terms of the mean and standard deviation of the RE for all the objects and samples}
\label{tab:experiment2}

\begin{tabular}{|c|c|c|c|c|}
\hline
                   &\textbf{Rigidity/Material} & \textbf{Size (mm)} & \textbf{Sensor 1 (nº samples)}& \textbf{Sensor 2 (nº samples)}\\ \hline
\textbf{Red cup} &Very High/Metal & 79x82x109 (Medium)& 0.083 $\pm$  0.120 (470)& 0.181 $\pm$  0.166 (346)\\ \hline
\textbf{Pringles bottle} &Medium/Cardboard & 230x74x74 (Big)&  0.165 $\pm$  0.210 (371)& 0.121 $\pm$ 0.238 (343)\\ \hline
\textbf{Green toy} &Medium/Plastic & 76x76x76 (Medium)& 0.151 $\pm$  0.214 (281)& 0.095 $\pm$  0.176 (314)\\ \hline
\textbf{Mustard bottle} &High/Plastic & 190x95x53 (Big)& 0.139 $\pm$  0.189 (317)& 0.153 $\pm$  0.144 (356)\\ \hline
\textbf{Soccer ball} &High/Rubber & 93x93x93 (Medium)& 0.068 $\pm$ 0.103 (387)& 0.101 $\pm$  0.123 (383)\\ \hline
\textbf{Rubik's cube} &High/Plastic & 56x56x56 (Small)& 0.130 $\pm$ 0.096 (562)& 0.127 $\pm$  0.189 (485)\\ \hline
\textbf{Peach} &High/Plastic & 56x62x59 (Small)& 0.094 $\pm$ 0.091 (712)& 0.108 $\pm$  0.136 (595)\\ \hline
\textbf{Strawberry} &High/Plastic & 53x44x41 (Small)& 0.123 $\pm$ 0.105 (457)& 0.090 $\pm$  0.094 (484)\\ \hline
\textbf{Dice} &Very High/Plastic & 16x16x16 (Very Small)& 0.174 $\pm$ 0.102 (371)& 0.109 $\pm$  0.187 (372)\\ \hline
\textbf{Marble} &Very High/Glass & 21x21x21 (Very Small)& 0.164 $\pm$ 0.140 (511)& 0.183 $\pm$ 0.174 (437)\\ \hline
\end{tabular}
\end{table*}

In order to analyze the behavior of our method with respect to the rigidity of the objects, we grouped them into three categories: \textit{Very High} (2,507 samples), \textit{High} (4,738 samples), and \textit{Medium} (1,309 samples). Our results showed that \textit{RGBmod} obtained a lower RE value (0.111 $\pm$ 0.1296) with the \textit{High} group, compared to the \textit{Medium} (0.134 $\pm$ 0.213) and \textit{Very High} (0.148 $\pm$ 0.155) groups. This means that our method estimates forces with a lower RE when grasping rigid objects than when grasping very rigid or slightly deformable objects. On the one hand, objects in the \textit{Very High} group tend to cause higher forces during contact because their surface is harder, which can saturate the sensors. On the other hand, deformable or slightly deformable objects, such as those in the \textit{Medium} group, generate less significant features in the tactile image during the grasping because they absorb part of the applied force, making the force estimation more difficult. 

In addition, we organized the objects in four groups depending on their dimensions: \textit{Very Small} (Dice, Marble, 1691 samples), \textit{Small} (Rubik's Cube, Peach, Strawberry, 3295 samples), \textit{Medium} (Red cup, Green toy, Soccer ball, 2181 samples), and \textit{Big} (Pringles, Mustard, 1387 samples). The RE values obtained with the \textit{Small} and \textit{Medium} groups were smaller (0.1106 $\pm$ 0.123 and 0.109 $\pm$ 0.155) compared to the \textit{Very Small} (0.159 $\pm$ 0.156) and \textit{Big} (0.145 $\pm$ 0.199) groups. We believe this is because \textit{Very Small} objects cover less surface area of the gel, resulting in a higher force at a single point. Also, the Dice and Marble objects are very rigid. In contrast, objects in the \textit{Big} group can have larger contact areas that cover almost the entire surface of the gel. In both cases, with \textit{Very Small} and \textit{Big} objects, the sensor can suffer a saturation that affects the force estimation.

\subsection{Experimentation with non-symmetrical objects}
\label{sec:non_symmetrical_objects}

As mentioned above, our force estimation methods are not limited to grasp symmetrical objects, but we must use symmetrical contact surfaces to ensure that the applied forces in both sensors (PapillArray and DIGIT) are very similar, so that we can properly compare the ground truth and predicted forces. Therefore, we can use our trained models to estimate grasping forces using only the DIGIT sensors in both fingertips.

\begin{figure}[htbp]
     \centering

     \begin{subfigure}[b]{0.23\textwidth}
         \centering
         \includegraphics[scale=0.24]{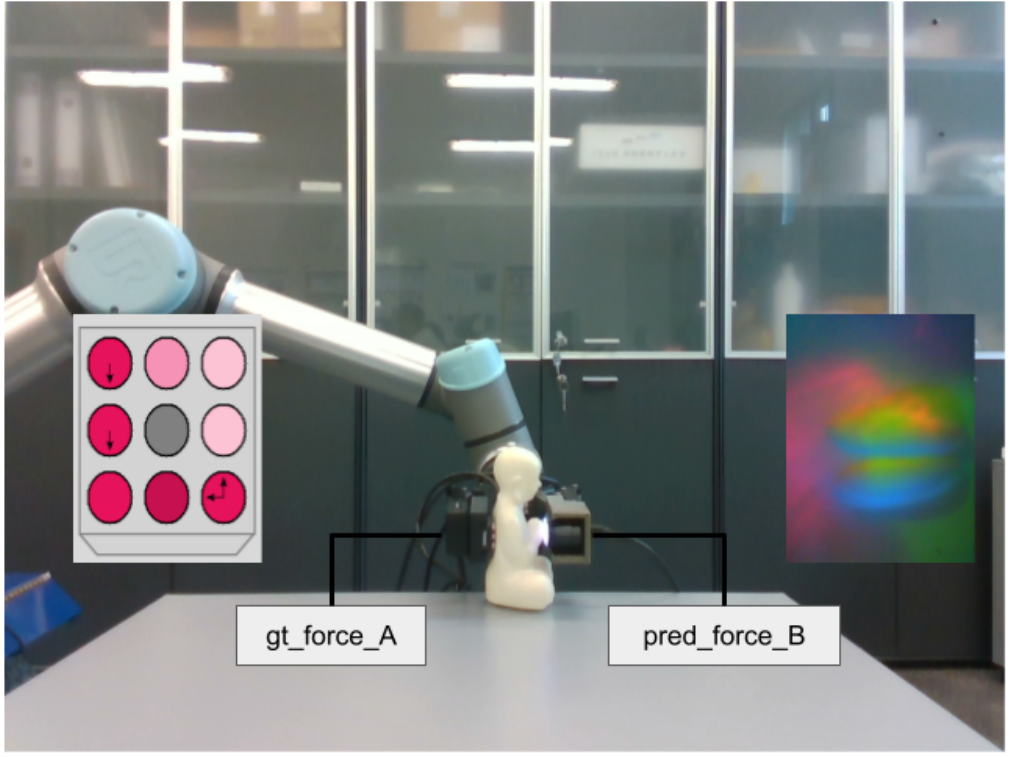}
     \end{subfigure}
    \begin{subfigure}[b]{0.23\textwidth}
         \centering
         \includegraphics[scale=0.24]{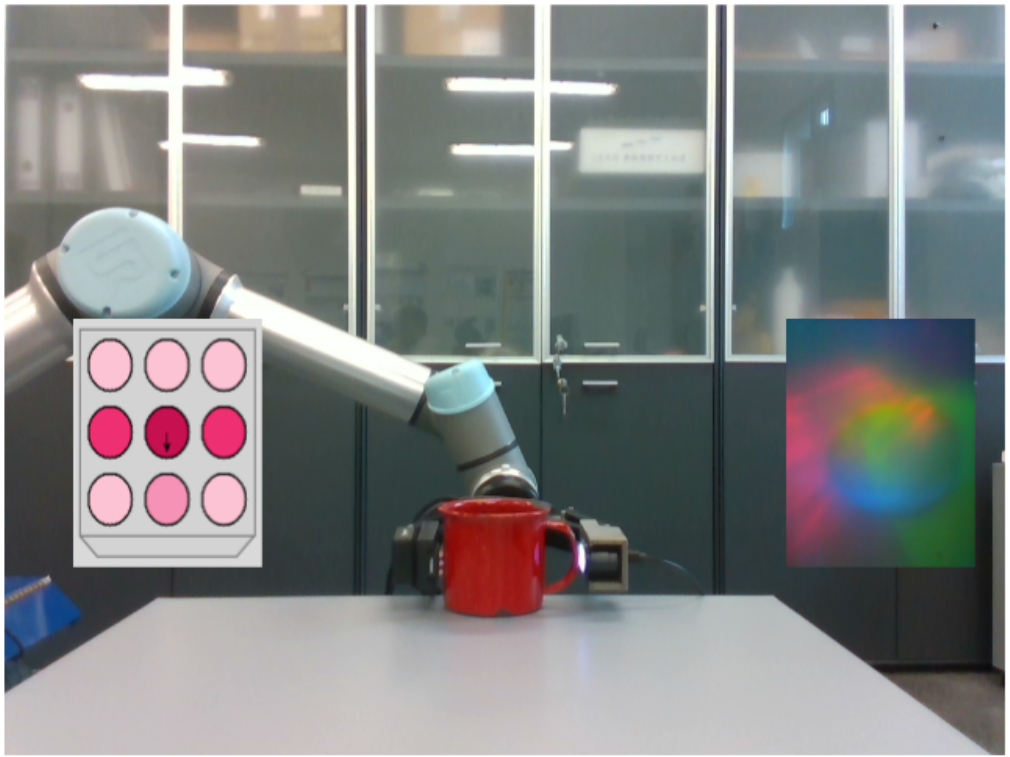}
     \end{subfigure}
     
      \begin{subfigure}[b]{0.23\textwidth}
         \centering
         \includegraphics[scale=0.24]{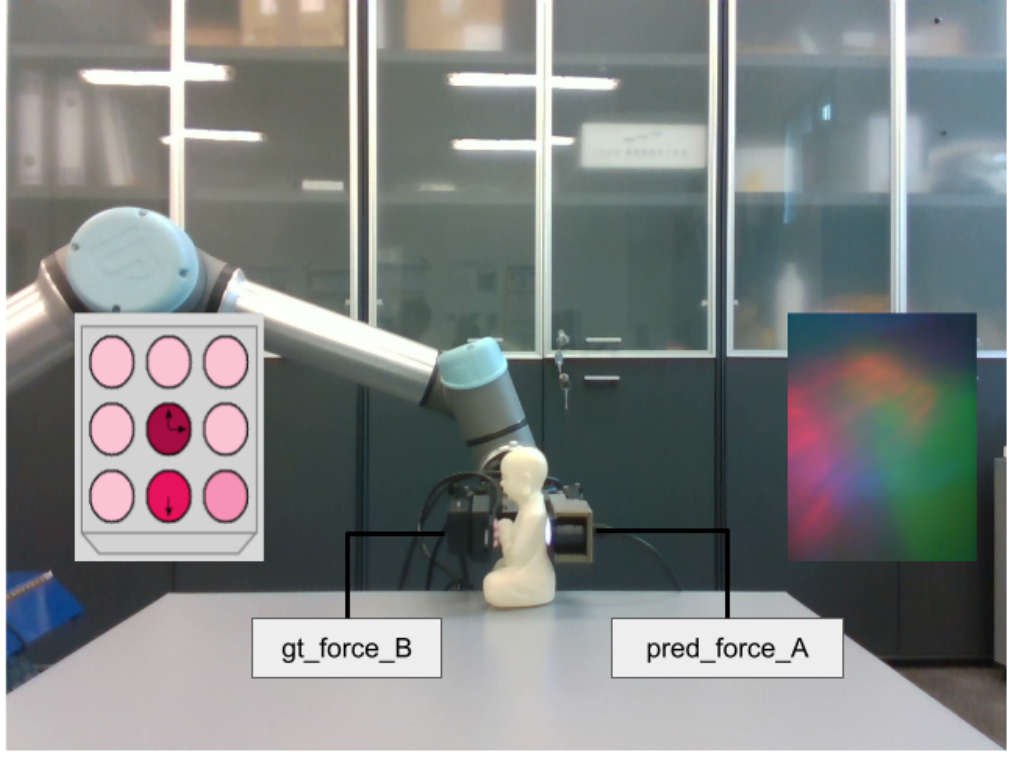}
     \end{subfigure}
     \begin{subfigure}[b]{0.23\textwidth}
         \centering
         \includegraphics[scale=0.24]{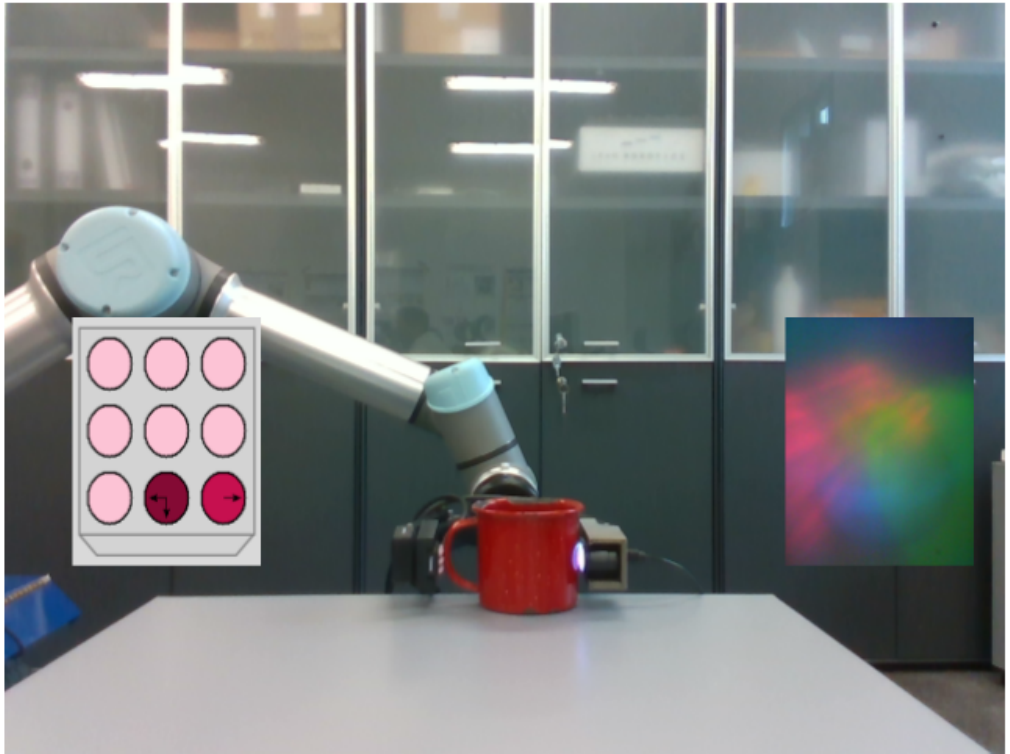}
     \end{subfigure}
    
    \begin{subfigure}[b]{0.23\textwidth}
         \centering
         \includegraphics[scale=0.28]{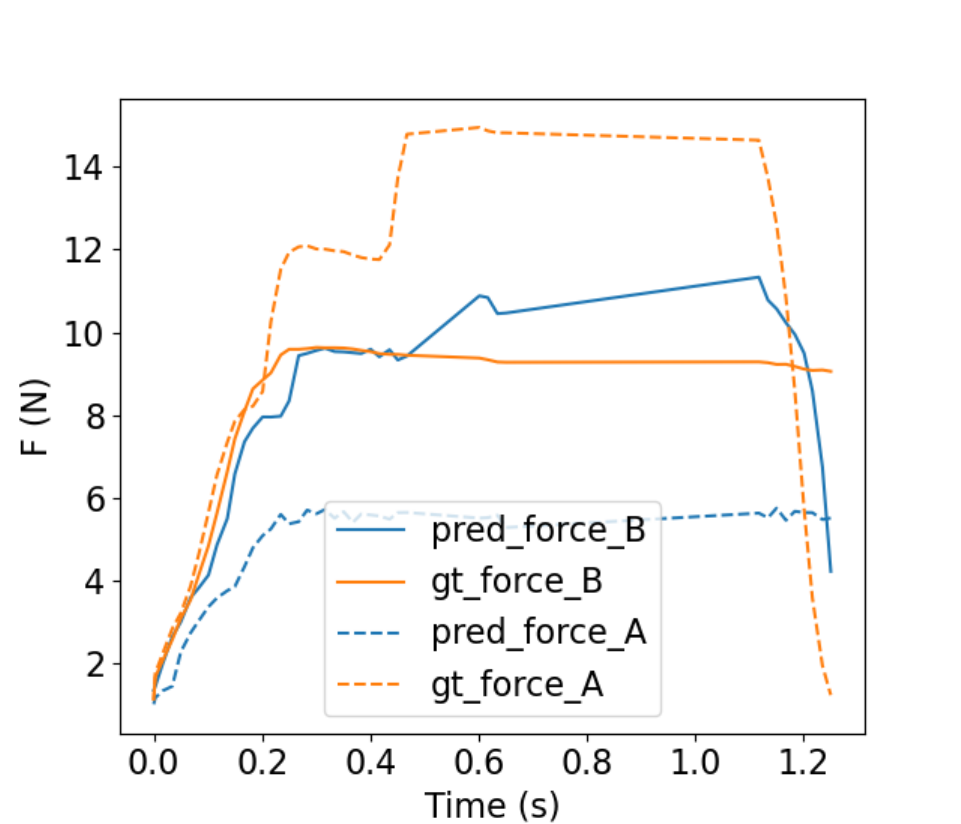}
     \end{subfigure}
      \begin{subfigure}[b]{0.23\textwidth}
         \centering
         \includegraphics[scale=0.28]{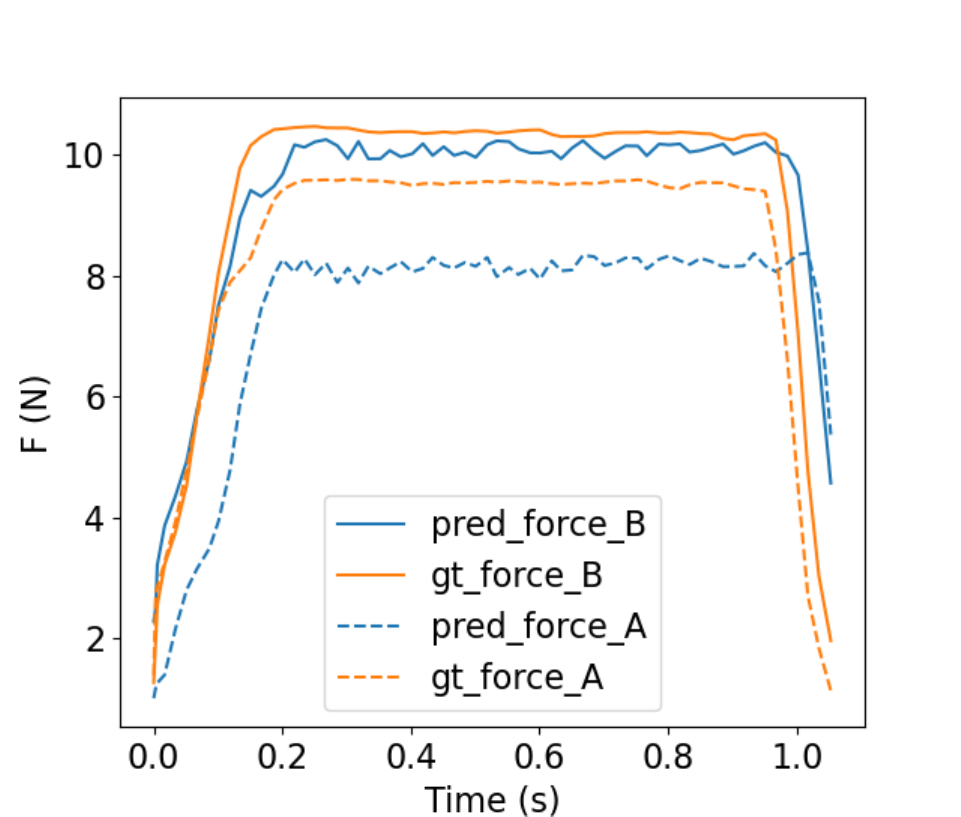}
     \end{subfigure}

      \caption{Examples of the estimated normal forces with \textit{RGBmod} when grasping unseen objects by non-symmetrical contact surfaces}
     \label{fig:non_symmetric_objects}
\end{figure}

Specifically, in this section, we performed an experiment on grasping non-symmetrical objects (White figurine, Red cup with handle, and Red pen). Note that in order to measure the force error with non-symmetrical contact surfaces, we need to grasp the object, flip the object or the gripper, repeat the grasp, and then cross the tactile data. Therefore, we calculated the force error between the predicted force of the first grasp and the ground truth force of the second grasp on the same contact surface of the object, and vice versa, as shown in Fig. \ref{fig:non_symmetric_objects}.

In contrast to the previous experiment with symmetrical objects, in this experiment with non-symmetrical contact surfaces we have measured the RE in the time interval between $t=0.2$ and $t=0.7s$, thus ensuring that the predicted and the ground truth forces contain the same number of samples.
After grasping three non-symmetrical objects a total of 18 times (510 samples), our method obtained an RE of 0.189 $\pm$ 0.215, which is 0.064 $\pm$ 0.062 higher than the average RE value obtained in the previous section. We believe that this decrease in performance depends on the type of contact surface. For example, the contact surface of one side of the White figurine is very small and pointed (pred\_force\_B, gt\_force\_B), and the other side is concave (pred\_force\_A, gt\_force\_A). Our method performed well on the very small and pointed side, but performed very poorly on the concave side because the tactile features were harder to detect (pred\_force\_A). On the other hand, for a non-symmetrical grasping of the Red cup or the Red pen, our method estimated the grasping force better on both sides. In addition, the rotation of the object or the gripper can also introduce some errors in the measurements.

\subsection{Comparison with other works from literature}

To conclude the experiments, we compared our methods with other state-of-the-art works that do not use visual markers for the estimation of grasping force \cite{shukrullo}, \cite{2}.

In \cite{shukrullo}, the authors calibrated a DIGIT sensor to obtain depth images following the same process as we did in Section \ref{sec:calib_process}, and fitted a third-order polynomial regression model to map a deformation value to the applied force in a teleoperated robotic grasping task. In order to compare it with our method, we evaluated their method on our dataset by obtaining the deformation values from the depth images for all the samples of the three datasets in Table \ref{tab:experiment1}.

Table \ref{tab:comparison1} shows the error comparison between our proposed approaches and our implementation of \cite{shukrullo} with the three datasets, since their code is not publicly available.
\begin{table}[htpb]
\centering
\caption{Comparison between the proposed approaches and \cite{shukrullo} for the three datasets in terms of the RE metric}
\label{tab:comparison1}

\begin{tabular}{|c|c|c|c|}
\hline
                & \textbf{S1}    & \textbf{S2}    & \textbf{S3}    \\ \hline
\textbf{RGBmod}    & 0.173 $\pm$ 0.306 & 0.206 $\pm$ 0.392 & 0.222 $\pm$ 0.447 \\ \hline
\textbf{D} & 0.911 $\pm$ 1.631& 0.949 $\pm$ 1.689& 0.920 $\pm$ 1.626\\ \hline
\textbf{RGBmod+D}      & 0.186 $\pm$ 0.279& 0.205 $\pm$ 0.389& 0.213 $\pm$ 0.433\\ \hline
\textbf{Dmod}      & 0.867 $\pm$ 1.501 & 0.919 $\pm$ 1.623 & 0.903 $\pm$ 1.603 \\ 
\hline
\textbf{\cite{shukrullo}}      & 0.576 $\pm$ 1.019 & 0.569 $\pm$ 0.984 & 0.576 $\pm$ 0.991 \\ \hline

\end{tabular}
\end{table}
These results lead us to the following conclusions. First, \cite{shukrullo} obtains lower RE values for the three datasets than our proposed approaches \textit{D} and \textit{Dmod}, meaning that a single deformation value is a better depth representation than a depth image for the task of force regression with the DIGIT sensors. This may be because CNNs have more difficulty extracting features from binary images than from color images. Second, our RGB-based methods outperformed \cite{shukrullo}, proving that a single deformation value is a worse tactile representation compared to RGB images for this task.

We also compared our approach \textit{RGBmod} with another state-of-the-art markerless method \cite{2}, which estimates a single force value distributed over the segmented contact patch. The method presented in \cite{2} was trained and tested using data from a Gelsight Mini sensor. We used their public code, trained model, and dataset of 277,325 images for the comparison. We trained our method using their dataset, and evaluated both methods using two versions of their test set: (i) a \textit{Full test set} containing ground truth forces between 0 and 20N, and (ii) a \textit{Filtered test set} containing forces between 1 and 15N. Both methods were evaluated using the RE and MAE metrics. Table \ref{tab:comparison2} shows a comparison of the results.

\begin{table}[htpb]
\centering
\caption{Comparison between our proposed method (\textit{RGBmod}) and \cite{2} for the two versions of the test set in terms of the MAE and the RE metrics}

\label{tab:comparison2}
\begin{tabular}{|l|ll|}
\hline
\textbf{}        & \multicolumn{2}{c|}{\textbf{Full test set}}                                     \\ \hline
\textbf{}        & \multicolumn{1}{c|}{\textbf{MAE (N)}}            & \multicolumn{1}{c|}{\textbf{RE}} \\ \hline
\textbf{\cite{2}} & \multicolumn{1}{l|}{0.796 $\pm$ 0.843}          & \textbf{0.205 $\pm$ 0.295}          \\ \hline
\textbf{Ours}  & \multicolumn{1}{l|}{\textbf{0.739 $\pm$ 0.650}} & 0.265 $\pm$ 0.493                   \\ \hline
                 & \multicolumn{2}{c|}{\textbf{Filtered test set}}                                 \\ \hline
                 & \multicolumn{1}{c|}{\textbf{MAE (N)}}            & \multicolumn{1}{c|}{\textbf{RE}} \\ \hline
\textbf{\cite{2}} & \multicolumn{1}{l|}{0.906 $\pm$ 0.821}          & 0.187 $\pm$ 0.131                   \\ \hline
\textbf{Ours}  & \multicolumn{1}{l|}{\textbf{0.703 $\pm$ 0.628}} & \textbf{0.168 $\pm$ 0.153}          \\ \hline
\end{tabular}
\end{table}

Our method achieves lower MAE values for both versions of the test set and a lower RE value for the \textit{Filtered test set}, while \cite{2} achieves a lower RE value for the \textit{Full test set}. This comparison leads us to conclude that both methods achieve a similar performance when evaluated with a range of forces between 0 and 20N, and that our method outperforms theirs when working with a more specific range of forces (1-15N). It is important to remember that these types of sensors tend to saturate at high forces (from 15N in the case of DIGIT). Grasping forces up to 15N are large enough to grasp and lift many different everyday objects. In addition, our method is faster in both the training and the inference time. Our method takes approximately 5 hours to train with their dataset and runs at 0.0009s, while their method takes about 22 hours for the training phase and runs at 0.006s. Although the difference in inference time may not seem particularly large, it becomes important when running these methods on embedded systems where the robot controller needs to be as fast as possible and other algorithms need to run simultaneously.

\section{Conclusions}
\label{sec:conclusions}

In this work, we have designed, implemented and evaluated several approaches for estimating grasping normal forces from different types of visuotactile data: (i) RGB without visual markers, (ii) depth, and (iii) a combination of both. Our code and dataset are available at this \href{https://aurova-projects.github.io/force_tactile/}{\textbf{link}}. 

After comparing the implemented approaches with our dataset, we found out that RGB tactile images are a better visuotactile representation than depth images for the task of force estimation with the DIGIT sensors. Moreover, we evaluated our best method, \textit{RGBmod}, by grasping different symmetrical and non-symmetrical unseen everyday objects in real cases. We obtained an average RE of \textbf{0.125 $\pm$ 0.153} for symmetrical objects and \textbf{0.189 $\pm$ 0.215} for non-symmetrical objects, which we consider as acceptable errors to perform stable grasps for different manipulation task such as pick and place. Finally, we demonstrated that \textit{RGBmod} outperformed other state-of-the-art works when tested under the same conditions, showing also that our method can estimate forces with different types of VBTS, i.e., the Gelsight Mini sensor.

A limitation of our proposal is that it performs worse when grasping objects with small (0-1N) or large (16-20N) forces. As explained in this work, small forces do not generate noticeable changes on sensor surfaces like DIGIT, and large forces tend to saturate the gel of the sensor. This can be solved by modifying the physical properties of the gel, but this is beyond the scope of this article. In addition, the performance of our method is dependent on the contact location on the gel, performing better when touching the center or top, and worse at the edges. However, this is because the DIGIT has a curved surface and the features in the tactile image are more difficult to detect in these areas. Note that in this work, we focus only on a single normal force, but we plan to extend our method to estimate multiple normal forces applied in different regions of contact.

\end{document}